# Model of human cognition

Wu Yonggang

October 22, 2025

# Table of Contents

# 1. Introduction

Recently, there has been immense development in the field of artificial intelligence (AI) and computational neuroscience. Numerous architectures and models have been implemented in artificial systems to challenge human intelligence, especially with the release of increasingly proficient large language models (LLMs). However, despite advancements in LLMs, artificial systems still fall short in matching the human capacity for generalisation across diverse tasks and environments, thus being an overstatement to label the current generations of LLMs as artificial general intelligence (AGI). We propose that in order to create artificial systems with high generalisation capabilities, one must first examine and understand the fundamentals of human cognition through conceptual models of the brain.

This paper introduces a theoretical model of cognition that integrates biological plausibility and functionality, encapsulating the fundamental elements of cognition and accounting for many psychological and behavioural regularities. The model consists of four main modules: the visual processing module, the semantic module, the predictive module, and the executive module. The modules are discussed in chronological order, with each being affiliated with corresponding anatomical regions of the brain. Thereafter, the model is substantiated with real-world examples and that reflect its general problem-solving capabilities.

# 2. Module 1: Visual processing system (VPS) module

Corresponding anatomical regions(s): V1, V2, V4, IT

The ventral stream consists of multiple hierarchical regions, beginning in the primary visual cortex (V1) and proceeding through V2 and V4 before reaching the inferior temporal (IT) cortex. Neurons in the ventral stream are feature-selective and exhibit preferred feature tuning. Across hierarchical regions, neurons are tuned to respond to different features, with more complex features being detected at higher hierarchical levels (Hochstein & Ahissar, 2002). V1 simple cells detect edges of different orientations and spatial frequency (Khan et al., 2011; Zhu et al., 2010). V2 neurons demonstrate selectivity towards contours, angles (Ito & Komatsu, 2004), and "T" junctions. V4 neurons detect simple shapes, longer and more complete contours (Tang et al., 2020). IT neurons demonstrate categorical invariance (Yamins et al., 2014) and can represent specific objects or object categories.

Neurons in the ventral stream possess tuning curves, where firing rate peaks for a preferred stimulus (like a perfect circle) and gradually diminishes for similar non-identical stimuli (like an oval) (Inman, 2006; Tang et al., 2020). A neuron also has a receptive field, or the visual region where stimuli resembling the tuned feature can fire the neuron (Li et al., 2016). Receptive field sizes generally increase across visual hierarchies (Desimone & Schein, 1987).

Lateral, feedforward, and feedback connections characterise neural connectivity in the VPS. Feedforward connections transmit information to deeper hierarchical layers, lateral connections link neurons within the same layer, and feedback connections carry top-down signals that bias neurons towards detecting task-relevant features. (Gilbert & Li, 2013)

## 2.1 Complementary plasticity hypothesis (CPH)

Unlike simple perceptron units, biological neurons are much more complex, with a single neuron capable of modelling the behavior of a deep neural network (Cazé et al., 2013). Precise spatial-temporal dynamics of input signals determine whether the neuron reaches its threshold potential and fires a spike (Margulis & Tang, 1998; Segev & London, 2000). Factors such as dendritic shunting, receptor saturation, non-linear dendritic integration, or excitation of inhibitory interneurons can give rise to complex input-output relationships (Zhang et al., 2013; Prescott & De Koninck, 2003).

We propose a general rule of neural integration, termed the complementary plasticity hypothesis (CPH), where combinations of presynaptic input firing patterns determine the strength of excitatory signal and the firing dynamics of the postsynaptic neuron.

Let a neuron $A$'s firing pattern be labelled as $A_n$, where $n$ is a firing pattern of a particular firing rate. Suppose that during the presentation of a visual stimulus, four presynaptic firing patterns with some firing rate, indicated below, are observed before the firing of the postsynaptic neuron.

$$A_9, B_5, C_6, D_5$$

In spike-timing-dependent plasticity (STDP), when a presynaptic spike consistently precedes a postsynaptic spike, long-term potentiation (LTP) occurs. Conversely, if the order is reversed, with postsynaptic spiking consistently preceding presynaptic spiking, LTD dominates (Feldman, 2012). Expanding upon the principles of STDP, we suggest that the postsynaptic neuron undergoes synaptic change to strengthen all possible combinations of presynaptic firing patterns, such that future encounters with these combinations will provide an additional excitatory signal to the postsynaptic neuron, resulting in a stronger postsynaptic response of higher firing rate. Each combination is termed a complementary input, and is denoted as a concatenation of neurons and their respective firing rates.

Strengthening of complementary inputs: $A_9B_5C_6D_5$ , $A_9B_5$ , $C_6D_5$ , $B_5C_6$ , $A_9D_5$, $A_9C_6$ , $B_5D_5$

Strengthening of singular inputs: $A_9, B_5, C_6, D_5$

Complementary inputs can have weights, which determine the magnitude of the additional excitatory signal. Complementary inputs that consistently precede postsynaptic firing will be strengthened, thereby increasing their weight. Conversely, complementary inputs that do not correlate with the postsynaptic neuron's firing will be weakened (Stent, 1973; Feldman, 2012), resulting in reduced weight. Different presynaptic firing patterns and complementary weights can produce a wide range of postsynaptic responses, creating nonlinear postsynaptic neuron input-output relationships. CPH predicts that neurons that fire in isolation, or with irregular firing patterns that do not correspond to complementary input with high weights, do not generate a sufficiently strong excitatory signal for postsynaptic firing. Changes to complementary input weights may be facilitated by mechanisms such as the remodelling of dendritic branches or the growth of dendritic spines on the postsynaptic neuron.

Neurons exhibit selectivity for particular features, where their firing rate correlates with how closely a stimulus resembles its preferred feature (Hansel & van Vreeswijk, 2012). As the strengthening of complementary inputs elicits a greater postsynaptic response, indicated by a firing pattern of a higher firing rate, a complementary input encodes a set of features and feature resemblance pairs, which, upon strengthening to a postsynaptic neuron, increases the resemblance to some composed feature. This process is abstract and inductive, reflecting a hierarchical buildup of complexity, where complex features are constituted from less complex features.

CPH has many functional advantages. Considering all non-redundant complement inputs maximises data efficiency while increasing the accuracy of single-trial weight updates by reducing the impact of noise. Regulatory mechanisms such as homeostatic plasticity bolster feature selectivity by preventing dominance of particular input complements, ensuring that a minimal number of input patterns is required for postsynaptic firing. Increased selectivity enhances learning by ensuring that a sufficient set of complementary inputs is activated for the postsynaptic neuron to fire, and for strengthening to occur. Consequently, future weight changes will occur only in response to statistically significant stimuli that visually align with features similar to a neuron's tuning.

CPH will be the primary neural plasticity rule used in this paper and will be discussed further in subsequent sections. The following section will explore the adaptation of neuronal tuning to the environment, tuning curves, and robustness of neuronal tuning.

## 2.2 Training and tuning

While intrinsic mechanisms may bias neural tuning, neurons are predominantly tuned via experience-driven plasticity (Karmarkar & Dan, 2006). According to CPH, a neuron is strengthened by frequently occurring complementary inputs that drive its firing. The neuron becomes increasingly responsive to the input firing pattern, firing more frequently across similar patterns. This repeated activation strengthens those connections, creating a positive feedback loop. Consequently, the neuron develops selectivity for the particular feature encoded by that firing pattern and becomes tuned to detect it. If the consistent input firing pattern encodes a rounded contour, the neuron is tuned to detect rounded contours.

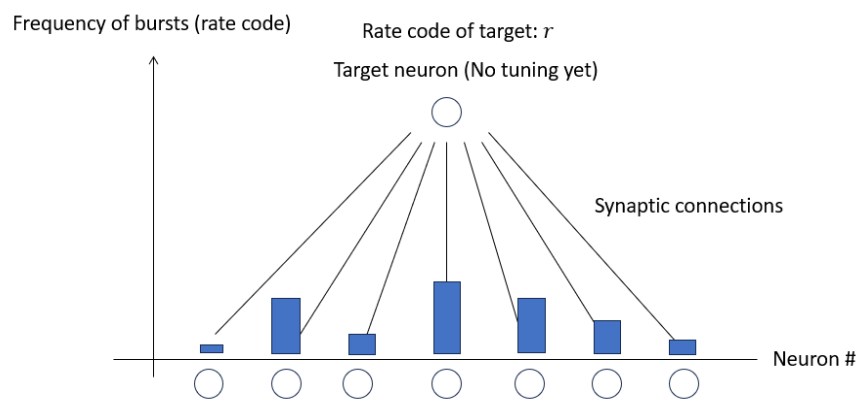

**Figure 2.1:** Illustration of a possible input firing pattern that a neuron may fire for. Assuming that the neuron has not been tuned yet, the feature encoded by the input firing pattern may become the neuron's tuned feature.

A drop off from the peak of the tuning curve signals a decline in feature resemblance to the complementary input's encoded feature. The weakened complementary input provides weaker excitatory signal, resulting in a weaker postsynaptic response. Tuning curves and complementary inputs provide intuition for the composition of features across hierarchical levels, where simple features at lower levels affect the recognizability of complex features at higher levels. For example, a line being oriented incorrectly may result in a contour being too flat, an ellipse being perceived instead of a circular wheel, which relinquishes the perception of a car. The relationship between complement inputs and tuning is further explored in **Figure 2.2**.

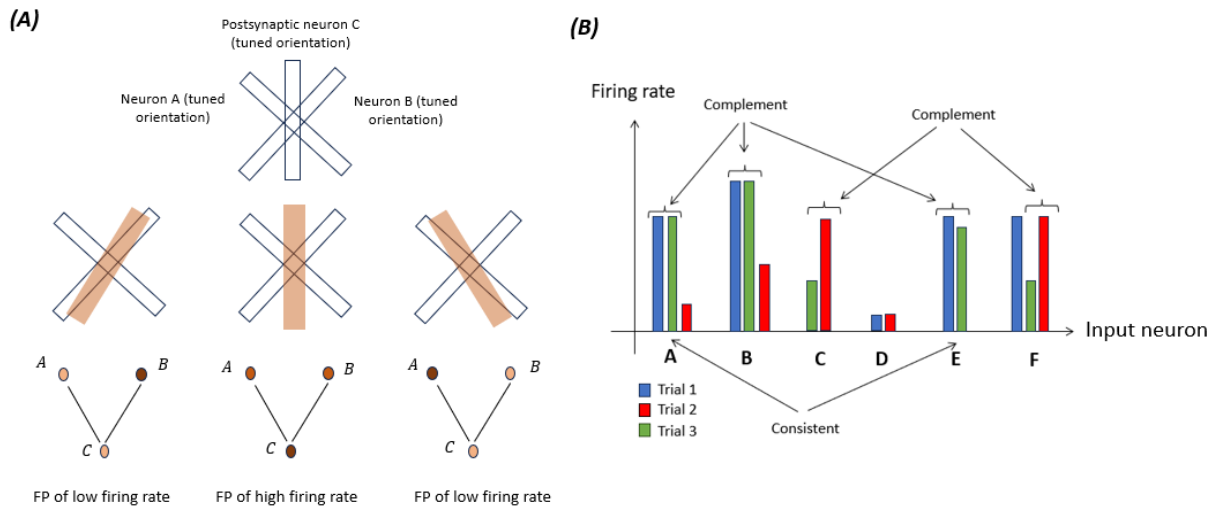

**Figure 2.2:** The postsynaptic neuron C is tuned to vertical bars of light. **(A)** Complementary inputs arise from combinations of firing patterns of particular firing rates. The figure illustrates how different complementary inputs between A and B can encode different features: bars of light of different orientations. **Left**: A combination of two firing patterns – B's firing pattern with a high firing rate, A's firing pattern with a low firing rate – encodes a bar of light tilted to its right. This complementary input elicits a weak response from C due to a deviation from C's preferred feature. **Centre**: The combination two firing patterns of A and B encode the feature of vertical bar of light, which matches the preferred tuning of C, eliciting a strong response in C. **Right**: A combination of two firing patterns – B's firing pattern with low firing rate, A's firing pattern with high firing rate – encodes a bar of light tilted to its left. This complementary input elicits a weak response from C due to a deviation from C's preferred feature. **(B)** An illustration of firing patterns of particular firing rates for six presynaptic neurons during each of the three perceptual trials, indicated by the colours of blue, red, and green. Input complements that are frequently observed across trials will be strengthened more to the postsynaptic neuron, resulting in those input complements developing stronger weights. Conversely, less frequently occurring input complements will develop weaker weights.

## 2.3 Specificity and invariance

In the primary visual cortex, neurons respond to a limited set of simple features, such as spatial frequency, translation, and orientation. The low dimensionality of the feature space in lower visual hierarchies results in overlapping and entangled representations rather than cleanly separable ones.

In higher visual areas, greater feature diversity increases the dimensionality of the feature space and the representational dissimilarity between stimuli. Features become more numerous and complex, demonstrating selectivity across details such as convexity (Pasupathy & Connor, 1999), boundary curvature (Kim et al., 2019), and texture (Okazawa et al., 2015). Dimension expansion allows for the disentanglement of stimuli by projecting them into a space where they can be linearly separated, a process necessary for object categorisation (Cohen et al., 2020). Dimension expansion also results in sparser activity, reducing the probability of neuronal firing during arbitrary natural image sequences.

The decline in neuronal firing probability means that synaptic adjustments occur less frequently, as postsynaptic firing is necessary for complementary weight strengthening. At higher visual areas, reduced synaptic adjustments and slower learning can potentially stall feedforward feature composition. However, further composition of features can still persist through the development of invariances. For example, a neuron that is invariant to position and scale can respond to a continuum of features observed at different sizes and positions, allowing it to fire with much higher probability than before. By conceptualising position and scale as dimensions within the feature space, a neuron that is invariant to position and scale can have its firing region span both dimensions in feature space, increasing the neuron's firing hypervolume.

Apart from invariances across simple transformations, invariances across convexity, viewpoints, and texture are also possible. For instance, face detecting neurons in IT demonstrate invariance for both profile and frontal views (Nam et al., 2021). Considering "profile" and "frontal" as separate dimensions within the feature space, invariance across both features can have a neuron's firing region span across both dimensions, increasing the neuron's firing theoretical hypervolume.

A balance between specificity, achieved through dimensional expansion, and invariance is critical for object recognition. Invariances support generalisations across similar features, while specificity amplifies differences among features for effective categorisation. Excessive invariance prevents proper decorrelation of features, causing neurons to fire too broadly across multiple features. Conversely, excessive specificity leads to neurons firing too infrequently, reducing the number of trials for strengthening complementary input weights and resulting in slow learning.

Nonetheless, invariance that occurs without a clear functional justification can be detrimental. Features that may appear visually similar should not be generalised automatically, as this can impair the ability to discriminate between objects. Conversely, visually distinct features should not be assumed to be unrelated, as different viewpoints of the same object may appear different yet belong to the same object category. We propose that invariances may be grounded upon two regularities of the physical world, namely:

1. Shorter temporal incidence between two visual stimuli is associated with a higher probability of both representing the same object or object category.
2. Features that persist over time tend to be more complex, abstract, and generalizable.

In the next section, a method of creating invariance through excitatory lateral connections will be proposed, attesting to the first physical world regularity.

## 2.4 Invariance through excitatory lateral connections

Biologically plausible models explain invariance through winner-takes-all (WTA) mechanisms (Nimmo & Mondragon, 2025), max pooling layers (Isik et al., 2012), or biological implementations of slow feature analysis (Lipshutz et al., 2020). This paper proposes another method of attaining invariance through excitatory lateral connections.

Invariance through excitatory lateral connections utilises the assumption that two stimuli occurring closely in time tend to encode varying reference points belonging to the same object. For example, ocular saccades, external forces acting on the object, or egocentric movement may create multiple reference points for the same object over time. Accordingly, invariance developed across visual stimuli occurring closely in time may gradually give rise to meaningful object categories through experience, thereby contributing to effective object categorisation. A step-by-step process for the development of invariances is proposed in **Figure 2.3**.

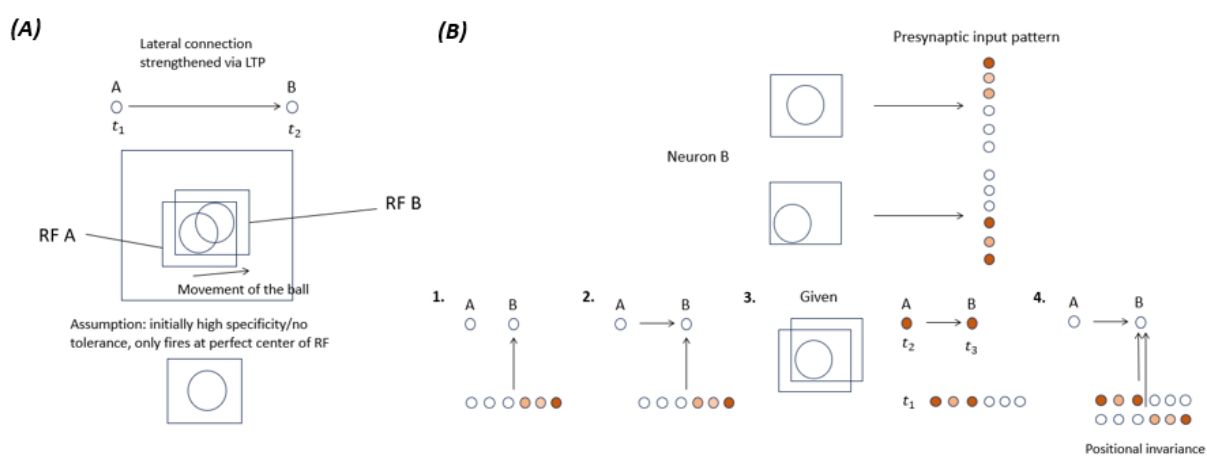

**Figure. 2.3:** Neurons A and B are tuned to circles. Initially, A and B fire only when the circle is at the centre of their receptive fields, with minimal positional invariance. We assume A and B have overlapping receptive fields, encompassing both receptive field centres, with sufficient space for additional movement of the circle. **(A)** When attending to a moving ball, the movement of the ball across space is reflected retinotopically, from the centre of neuron A's receptive field to the centre of neuron B's receptive field. This results in the sequential firing of A followed shortly by B, strengthening excitatory lateral connections between A and B. **(B)** An illustration of the development of positionally invariant neurons. Suppose that A and B each have six presynaptic inputs depicted as coloured circles, with firing patterns of particular rates corresponding to the colour intensity. **(1)** Firing patterns of input neurons when the ball appears at the centre of neuron A and B's receptive fields. **(2)** As the ball moves, sequential firing of A and B strengthens lateral connections between A and B, until they are sufficiently strong to support bidirectional firing. Thereafter, when the ball appears in A's receptive field centre, A will fire B. **(3)** The input pattern for B, when the ball appears in A's receptive field centre, is strengthened to B. B is now able to fire when the ball is at two positions. With multiple neurons being tuned to circles, bidirectional strengthening enables the ball to fire at multiple positions within their receptive fields, creating positionally invariant neurons.

Repeated coincident activation of two neurons in overlapping temporal windows can create strengthened bidirectional excitatory connections between them (Bi & Poo, 1998; Song et al., 2005), serving as the structural substrate for a cell assembly. With more "circle" responsive neurons sharing overlapping receptive fields, repeated bidirectional strengthening leads to the formation of a cell assembly. Neurons within the cell assembly exhibit positional invariance, since the activation of any neuron whose receptive field encompasses the ball activates all neurons within the assembly through strong reciprocal connections. Each neuron strengthens its input complements, encoding the relative positions of stimuli within its receptive field. In the visual cortex, such cell assemblies may manifest as cortical columns, representing clusters of functionally related neurons with overlapping receptive fields.

**Figure 2.3** assumes that a neuron fires only when the stimulus lies precisely at the centre of its receptive field. However, representing every possible position of a ball within a receptive field would theoretically require infinite neurons. In practice, neurons exhibit some degree of positional invariance arising from the progressive buildup of invariance, originating from simple to complex cell pooling in V1 (Lian et al., 2021). Initial positional invariances, together with multiple neurons possessing overlapping receptive fields encoding the same feature, ensure that all possible positions of the feature within the receptive field are accounted for.

## 2.5 Decorrelation through inhibitory lateral connections

Inhibitory interneurons enhance neuronal selectivity by decorrelating activity between neurons tuned to features belonging to different objects (Sippy & Yuste, 2013). While excitatory connections are responsible for the creation of cell assemblies that encode invariant object representations across different reference points, inhibitory connections serve a complementary role, supporting the formation of cell assemblies by decorrelating assemblies encoding features of different objects. A demonstration of the complementary role of inhibitory lateral connections is shown in **Figure 2.4A**.

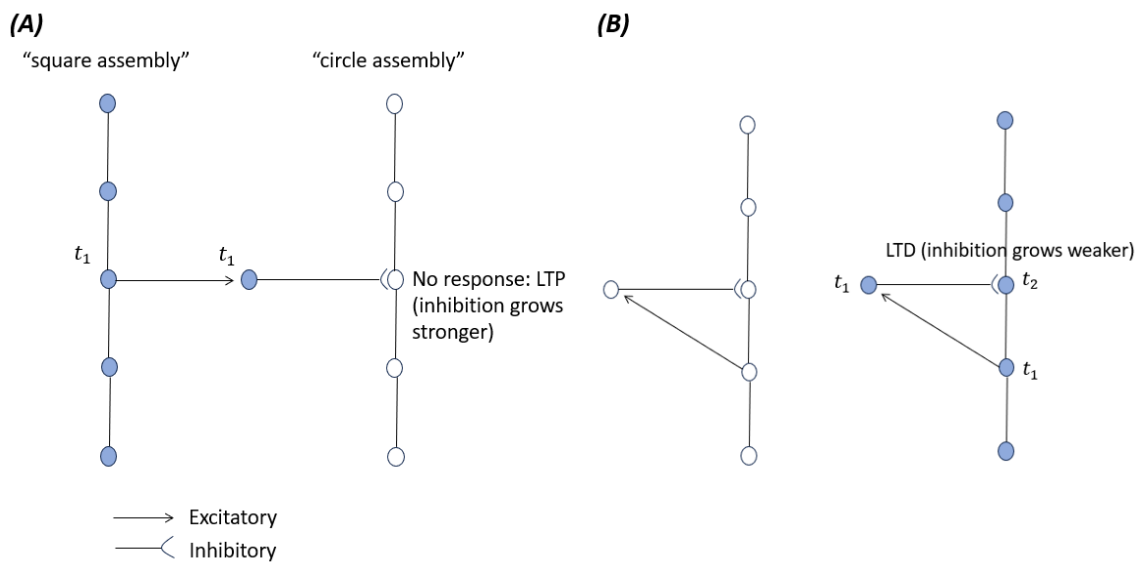

**Figure 2.4: (A)** An illustration of how inhibitory neurons mediate the decorrelation of assemblies. When a neuron within the "square" fires, it activates an inhibitory interneuron targeting a neuron in the "circle" assembly. Since squares do not transform into circles regardless of reference point changes, the firing of the "square" assembly rarely coincides with that of the "circle" assembly. An absence of postsynaptic activity induces LTP, consistent with the general anti-Hebbian STDP observed at inhibitory synapses (Kleberg et al., 2014) (Vogels et al., 2013). This strengthens the inhibitory suppression of the postsynaptic "circle" neuron, decorrelating activity between "square" and "circle" assemblies **(B).** In contrast, when the inhibitory neuron targets another neuron in the "square" assembly, its firing coincides with postsynaptic activity, resulting in LTD and weakening future inhibitory suppression of the postsynaptic "square" neuron.

Competition between separate assemblies can also promote feature selectivity, as neuron populations within different assemblies compete to recruit other neurons while preventing them from joining other assemblies. **Figure 2.6** illustrates a simple competition circuit, where inhibitory interneurons regulate the decorrelation of assemblies and the assignment of neurons to specific assemblies.

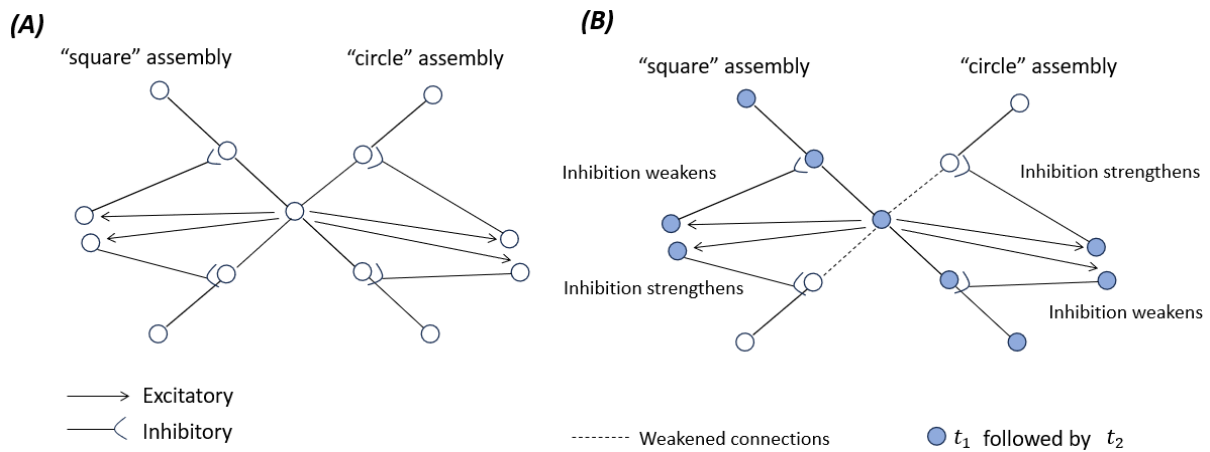

**Figure 2.5: (A)** An illustration of competition between the "square" and "circle" assembly, where each assembly compete to recruit the central neuron into its assembly. **(B)** Suppose that squares are observed more frequently than circles in the physical world. As the "square" assembly fires, the absence of a postsynaptic response in the "circle" assembly leads to LTP, strengthening inhibitory suppression of neurons in the "circle" assembly, decoupling the centre neuron from the "circle" assembly. Over time, the central neuron no longer fires for circles, shifting its tuning from circles to squares.

## 2.6 Complex invariances

Invariances in higher visual areas are more complex. Invariances to objects that are partially occluded, of different poses, viewpoints, or possessing deformations, cannot be created easily through excitatory lateral connections.

Invariance through excitatory lateral connections requires two conditions: an adequate set of reference points observed and sufficient neurons to encode them. For simple rotational, scalar and translational invariances, movement of the object or egocentric movements can provide many reference points to create invariant representations. Simple objects also tend to have a higher frequency of observations - any ball, wheel, or dot could fire a circle encoding neuron.

The development of viewpoint-invariant representations for complex, multifaceted objects would require multiple reference points from different angles to encode their complete 3D structure. For instance, an individual may have observed a giraffe standing upright, but never in its drinking posture, preventing a giraffe encoding neuron from developing viewpoint invariance. Additionally, objects rarely perform complete, full-view rotations within the precise temporal window required for effective STDP. Therefore, complex objects often lack sufficient reference points and encoding neurons in higher visual areas due to sparsity, hindering the formation of invariances through excitatory lateral connections.

Instead, complex invariances are usually created through the extraction of slow features – 2D statistical regularities that persist across multiple viewpoints. For example, a natural image sequence of a face may span a continuum from profile to frontal views. The fast-varying features correspond to continuously changing visual form of the face, including lines, curvatures and contours from front to profile view. Conversely, slow features represent stable properties, such as the presence of a nose, eyes, ears or skin texture, which can be generalised across multiple viewpoints. Complementary inputs encoding slow, stable features are activated and strengthened more frequently than inconsistent fast features, developing a higher complementary input weight. As a result, slow features exert more influence on postsynaptic firing than fast features, effectively approximating slow feature analysis (SFA).

## 2.7 Complex invariances and specificity

IT neurons do not treat an object as a holistic template but rather as a combination of latent features that integrate both form and spatial information. A cat's slow features may include its head, body, tail and legs, along with a generic spatial arrangement in which the head is connected to the body and the legs positioned beneath it.

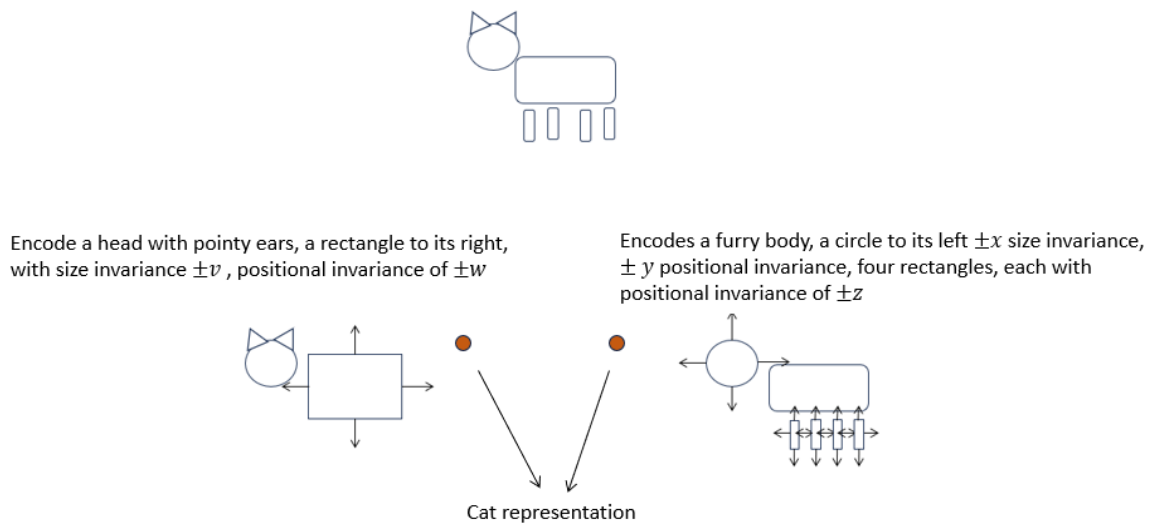

**Figure 2.6:** This diagram provides a simplified model of how neurons A and B leverage the principles of invariance and specificity to encode a cat. **(Left)** Neuron A encodes a conjunction of features with differing invariances: a head with cat ears, exhibiting minimal positional invariance, and a rectangular shape, exhibiting high positional and size invariance. **(Right)** Neuron B encodes another feature conjunction: a rectangular body with fur, a circle exhibiting

high positional and size invariance on its left, and a set of four rectangles below it exhibiting positional and size invariance. Each neuron represents a fragment of the cat; their joint activation provides a more robust and accurate representation.

As mentioned previously, a balance between invariance and specificity is critical for object recognition. Features cannot be too specific such that they can never be observed, yet cannot be too invariant such that feature composition no longer increases feature complexity. This balance can be attained by encoding a combination of features with different degrees of invariance and specificity. For instance, neurons selective to bicycles may encode the combination of a bicycle's handlebars (fast feature) with a rectangular shape representing the bicycle frame (slow feature). Or a bicycle wheel with low positional invariance (fast feature), with a bicycle seat with high positional invariance (slow feature). **Figure 2.6** illustrates another example involving the encoding of features with differing degrees of invariance and specificity.

## 2.8 Object categorisation

Single neurons encoding separate features of an object may not provide sufficient information to fully represent its identity. The complete representation of an object can be achieved through further hierarchical processing – either by postsynaptic integration of complementary inputs that encode multiple pieces of information, or through population coding supported by sparse cell assemblies. Cell assemblies provide functional redundancy as the combined activity of multiple neurons compensates for single neuron inconsistencies.

While further hierarchical processing with CPH could, in theory, give rise to "grandmother" cells – cells that respond only to a very specific concept or object (Gross, 2002) - the specificity of encoded representations is ultimately constrained by the frequency with which the object is encountered. For example, a neuron may become selective to a person A's face, but if person A is never seen again, the neuron will never fire. Over time, homeostatic plasticity and synaptic scaling suppress strong complementary inputs and strengthen weak complementary inputs, enabling the neuron to respond to less specific categories. A neuron's tuning may change from "face of person A" to a broader category, such as "face of male human", if person A is never seen again. This process prevents neurons from encoding rare exemplars that may not be useful for object recognition, with studies supporting that broadly tuned neurons have a greater impact on accurate object categorisation than neurons tuned to specific objects (Thomas et al., 2001).

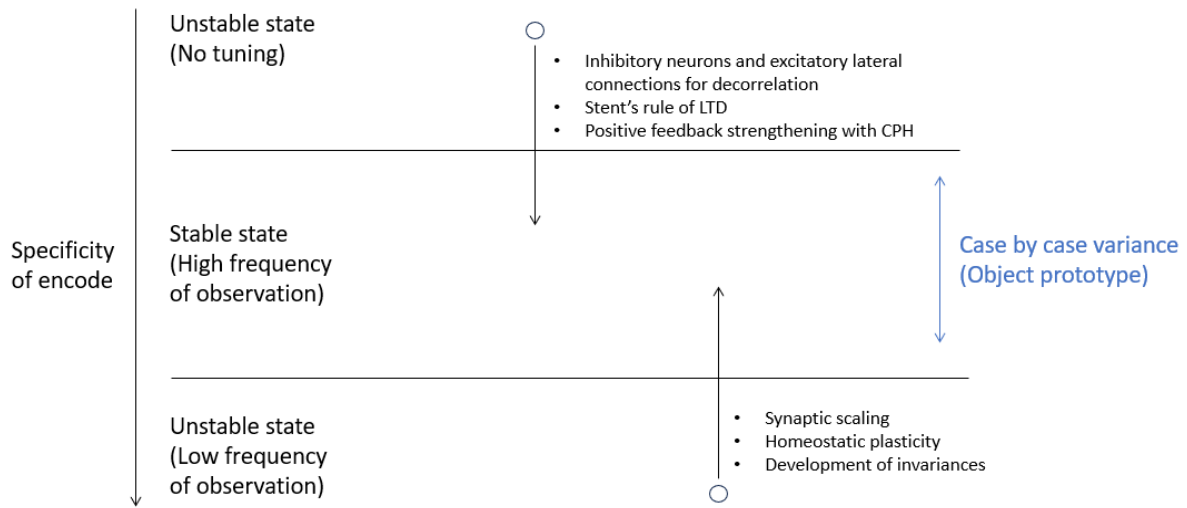

**Figure 2.7:** An illustration of neuron development that balances specificity and invariances. Neurons not tuned to any feature or that possess excessively specific tunings reside in unstable states. Neurons in unstable states converge towards stable states, where neurons are tuned to a specific feature that is observed frequently enough to complementary input weights. Reasons for convergence into stable states are annotated in the diagram. Within stable states, slight variations in the objects due to noise may result in different combinations of input firing patterns observed before postsynaptic firing. The statistical frequency of idiosyncratic variations is encoded by complementary input weights, forming the object prototype; specifics of object prototypes are discussed in the following module.

## 2.7 Object-specific synaptic adjustment

CPH suggests that consistent input patterns are strengthened during postsynaptic firing, while inconsistent input patterns are weakened. The important assumption is that strengthening of input complements only occurs upon postsynaptic neuron firing.

For postsynaptic firing, at least some features encoded by complementary inputs must be visually present. This implies that the visual stimulus necessary to elicit strengthening of input complements must share some features with the visual stimulus that the neuron was tuned for. If a neuron is tuned to detect faces, strengthening of complementary inputs occurs only when faces or non-facial objects resembling faces are presented (Bardon et al., 2022). Ensuring that strengthening is selective towards visual stimuli with similar features increases the likelihood of extracting object-relevant information. Additionally, spontaneous input activity would not provide a sufficient excitatory signal to induce postsynaptic firing, reducing the likelihood of random strengthening.

## 2.8 Development of visual neurons

During infancy, neurons in the visual cortex adapt to the environment by tuning to frequently occurring visual features (White et al., 2001; Ishikawa et al., 2018). Initially, a neuron's firing is driven either by arbitrary connections or intrinsic biases within inputs (Feller,

1999). This initial selectivity can be modelled as firing regions in a multidimensional space of input configurations (**Figure 2.8A**). As a toddler gains more experience, inhibitory decorrelation coupled with the preferential strengthening of a particular set of frequently encountered complementary inputs leads to the development of feature-selective neuron tunings (**Figure 2.8B**).

Excitatory lateral connections and synchronised neuron activity for neurons tuned to features belonging to the same object enable the development of cell assemblies, encoding an object representation that is invariant across simple details, and robust to idiosyncratic variability. Neurons within the assembly develop multiple input firing patterns, as shown in **Figure 2.8C**. Conversely, complex invariances are naturally created through neural tuning to slow features. While this does not create more input firing patterns, complex invariances are implicitly incorporated into a neuron's firing region through preferential strengthening of more frequently occurring input firing patterns. The neuron's tuned firing region in **Figure 2.8B** thus encodes the complex invariances.

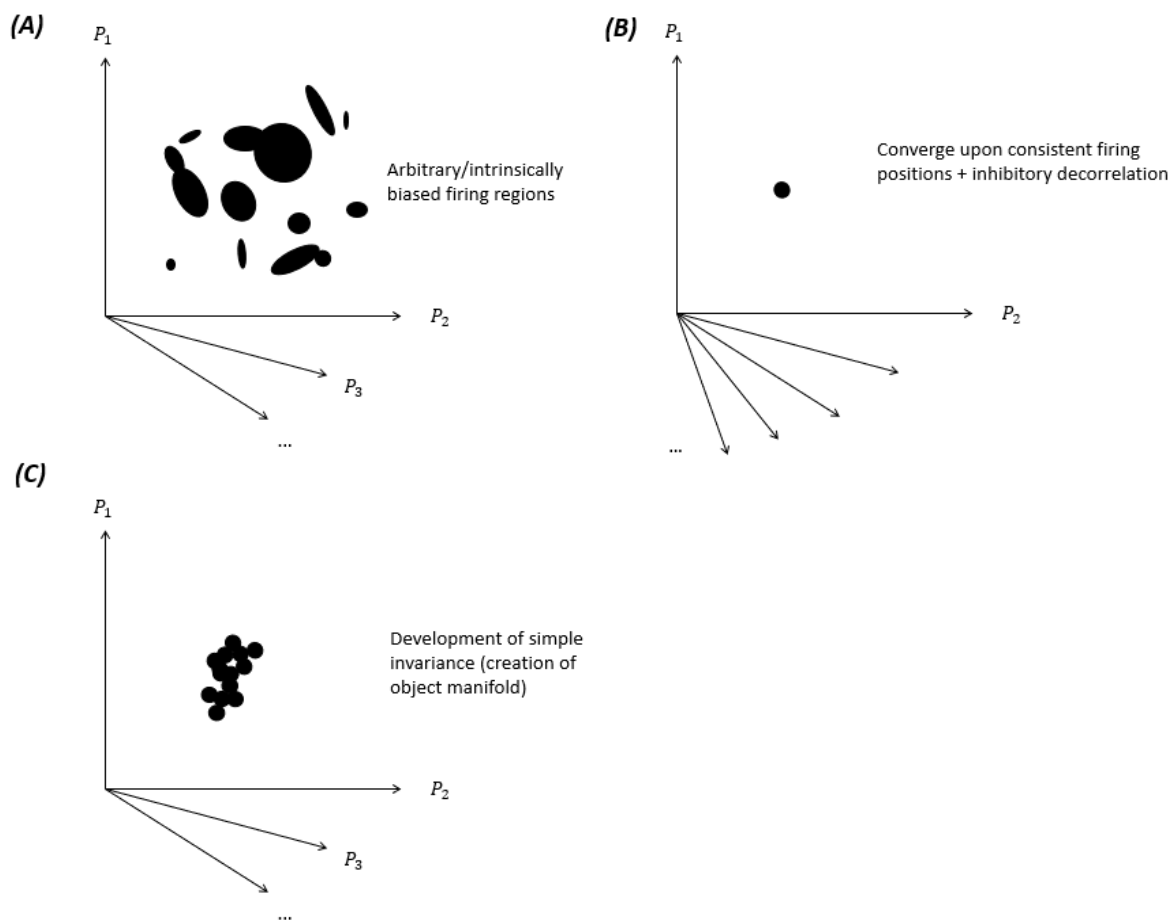

**Figure 2.8:** Illustration of a 3-step process for the development of a neuron in the visual processing module. The $P_n$ The Axis represents a linearised continuum of firing patterns for input neuron $n$. **(A)** Initially, the firing state space of a neuron has numerous arbitrary or intrinsically biased regions which can result in firing. **(B)** Over time, the number of firing regions is reduced due to neural tuning, resulting in a single firing region remaining, encoding a feature with a tuning curve represented by the size of the region **(C).** The creation of simple invariances leads to the proliferation of multiple firing regions, each representing different stimuli along an imaginary invariance continuum.

# 3. Module 2: Semantic hub module

Corresponding anatomical region(s): Anterior temporal lobe (ATL)

The anterior temporal lobe (ATL) integrates complex visual information through convergent projections from the IT cortex. Serving as a transmodal semantic hub, the ATL integrates information from different sensory modalities to form complex, high-level representations of objects (Hoffman et al., 2014). For simplicity, the precise spatial-temporal dynamics of input firing patterns will not be considered from section 3.1 onwards. Neuron activity will be treated as binary, either activated or not activated.

ATL utilises the combined activity of many neurons to represent objects through population codes. Population codes can provide a graded estimation of confidence (Bays, 2016). Suppose 100 neurons fire for dogs and 100 neurons fire for cats, with 10 overlapping neurons firing for both cats and dogs. When the visual stimulus of a dog morphs into a cat, dog neurons decline from 100 to 10, while cat neurons increase from 10 to 100. The point where an equal number of neurons encoding cats and dogs fire can represent a high degree of uncertainty.

To reiterate, population codes are robust to noisy and ambiguous stimuli, as object identity can be determined by population vote (Tanabe, 2013). For example, if there were only one neuron encoding a feature present in cars, a car with a dent may not contain that specific feature required for the neuron to fire. With multiple neurons corresponding to different features observed in cars, a car with a dent may disrupt the activity of a subset of car-feature neurons; the car representation may still be reflected in the other car-feature neurons.

Additionally, if the presence of a dent weakens a car feature neuron's firing, it may be encoding a latent feature that is present specifically in cars but not in dents. Conversely, if a car feature neuron fires more strongly due to a dent, it may encode a latent feature shared by both cars and dents.

The sparsity of neurons firing for a single object, coupled with the reuse of latent features shared across multiple objects, means that any stimulus, including novel, unseen stimuli, can be represented by some population of neurons (Haxby et al., 2001). Generally, the more similar two objects are, the more latent features they share and the more overlapping neurons they activate (Hebscher et al., 2023).

Neurons do not adopt holistic, fixed object templates; instead, they map out statistical regularities in the environment, not necessarily belonging to any specific semantic or object category (Ponce et al., 2019). Choosing or recombining different populations of neurons can amalgamate unique stimulus that an individual may not have encountered before (Reddy & Kanwisher, 2006). Visual categorisation through abstract neurons, leveraging statistical regularities, allows for novel object representations to be easily stored as reusable combinations of neurons, without requiring any change to the physiological neural structure. Examples of reusable abstract features, accompanied by the physical world objects that may contain those features, have been proposed for consideration:

Cars, apples, flamingos – likely a curved shape latent feature

All vehicles with wheels – likely a wheel + chassis latent feature

Cars, vaults, phones – likely a rectangular + metallic latent feature

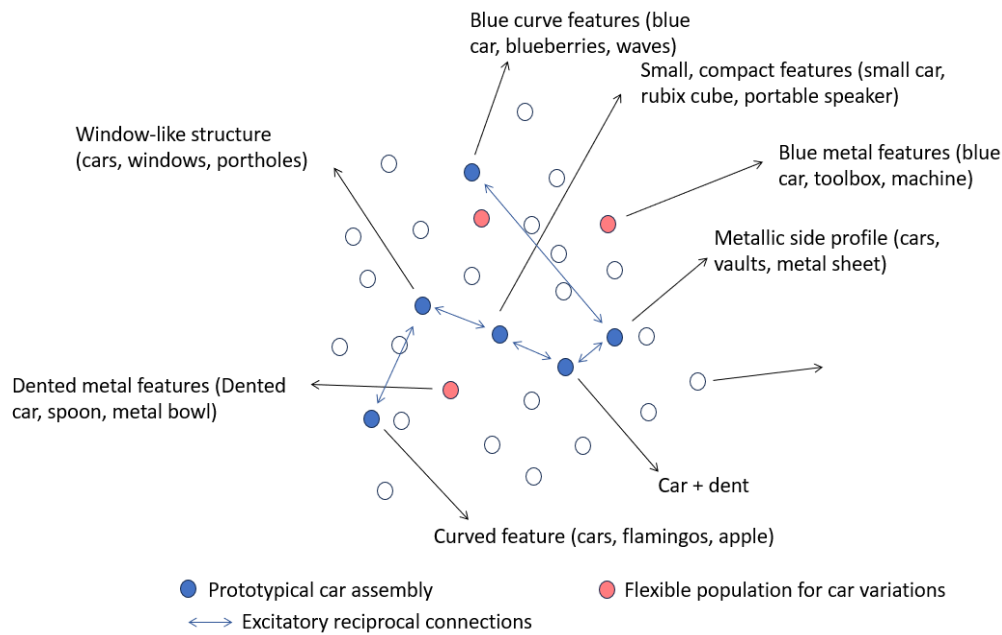

**Figure 3.1:** Neurons within the ATL receiving projections from the IT. Each neuron encodes different features, with each feature being present within multiple objects. Variability in idiosyncratic observations can cause neurons outside the typical assembly to fire, or prevent neurons within the typical assembly from firing.

## 3.1 Hub and spoke model

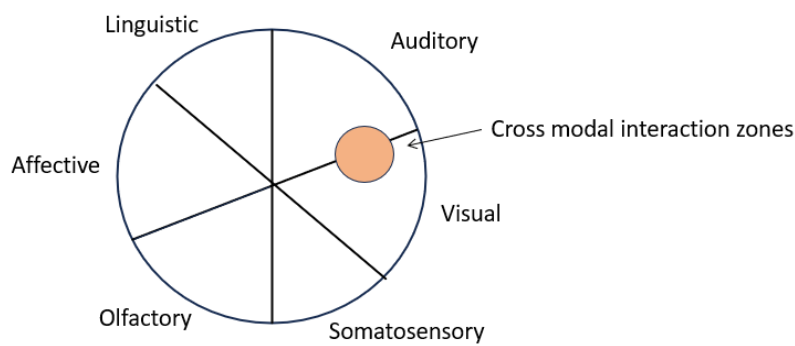

**Figure 3.2:** An illustration of the graded semantic ATL hub, with areas between two modality-specific regions of the ATL representing integration zones, where stimuli embody both modal qualities.

According to hub-and-spoke models of semantic memory, the ATL serves as the transmodal hub, and connections between modal-specific cortical areas and the ATL are the spokes (Patterson & Ralph, 2016). The ATL is not responsible for memory storage; rather, it serves as a binder and retriever of memory traces within cortical areas. Lesions to the ATL impair semantic access but do not erase its representations in cortical areas (Anand et al., 2025). A unified concept of a "car", for example, arises from the integration of features stored within distributed, modality-specific spoke regions. These include auditory areas for the sound of a car engine, visual areas for visual form, and language areas for word form.

The graded semantic hub hypothesis (Rice et al., 2015) extends the hub-and-spoke model, in which the topological distance from input connections to visual, auditory, or linguistic areas influences the relative strength and contribution of each modality within the representation. The functional properties of subregions depend on connectivity with spokes of different modalities, with peripheral areas being more modality-specific and central regions more integrative and a-modal (Binney et al., 2016; Rice et al., 2015).

For each spoke, the ATL receives preferential inputs from high-level sensory areas, such as the fusiform gyrus and IT for vision, and the superior temporal gyrus for audition (Papinutto et al., 2016), and receives more projections from cortical regions of closer proximity (Lv et al., 2023). Similarly, during retrieval of unimodal stimuli, feedback projections from the ATL terminate in mid- to high-level sensory areas, with no evidence of direct projections to primary visual areas (Rockland & Van Hoesen, 1994).

The ATL is a transmodal hub involved in the processing of conceptual information. Previous terms such as "features" or "objects" are more applicable to stimulus-specific characteristics but less appropriate for describing concepts. We introduce the standardized term "percept" to describe any representation in general, encompassing the totality of stimulus, features, objects, concepts, events, etc. A percept will be denoted by its lexical descriptor, enclosed in angle brackets; for instance, the visual form of a car will be <car (vis)>. Additionally, a collection of percepts at some point in time will be defined as a percept context.

## 3.2 Encoding of percepts

The ATL receives polysynaptic projections from the hippocampus via the entorhinal and perirhinal cortex, facilitating the transmission of hippocampal sharp wave ripples (SWRs) that replay a temporally compressed sequence of percepts for memory consolidation (Joo & Frank, 2018). Repeated replay of percepts within the spike-timing window through SWRs facilitates plasticity (Sadowski et al., 2011); backwards and forward replay of SWRs also allows for bidirectional association of percepts. (Diba & Buzsáki, 2007)

Learning-induced neural plasticity in the ATL and the VPS differs in its implementation. In the VPS, coincidental, memoryless firing of complementary inputs during online perception is the primary driver of plasticity, supporting intra-object recognition. In the ATL, SWRs encode sequences of objects over time, enabling associations to develop between multiple objects, reflecting the ATL's role as an inter-object associator.

## 3.3 Hetero-associations (One-for-one)

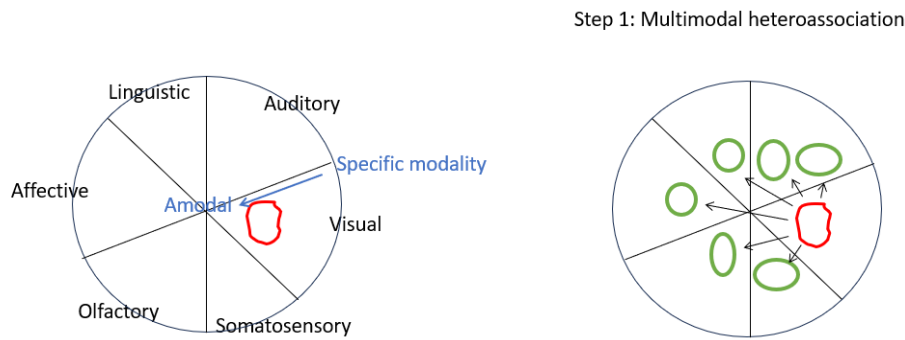

**Figure 3.3: (Left)** Proximity to the centre of the graded ATL hub corresponding to the increase in a-modality of a percept. **(Right)** Hetero-association of a visual object to possible conceptually related counterparts in other modalities.

We suggest that a transmodal, holistic conceptual representation of the percept can be developed through hetero-associations, in which a cell assembly encoding a percept can activate other assemblies encoding percepts of conceptual relevance through spreading activation. One-for-one hetero-association may be viewed as automatic retrieval: the real-time processing of percepts by sensory cortices automatically retrieves the corresponding holistic representation in the ATL.

To encode a relation between two percepts, the percepts need to be retrieved with some temporal difference, for instance, by observing the visual stimulus of a car before the written word of car. This sequence is encoded by the hippocampus, which obligatorily encodes all information that is attended to, whether internally through conscious apprehension or externally through perception (Aly & Turk-Browne, 2017; Nadel & Moscovitch, 1997). Serving as a fast-learning system, the hippocampus can encode episodic traces through one-shot learning (Sugar & Moser, 2019), which is then sequentially replayed to the cortex during sleep via SWRs, preserving either the natural or the reversed order during retrieval (Ecker et al., 2022). Repeated replay of compressed percept sequences provides precise spike sequences within brief windows, conducive for synaptic plasticity (Sadowski et al., 2011), effectively strengthening the connections between two assemblies and relating the percepts together.

We hypothesize that SWRs allow for synaptic strengthening to be conducted in a precise and targeted manner, minimizing interference with other existing relationships stored in orthogonal neural subspaces. Moreover, monosynaptic connections between neurons in ATL provide a direct connective pathway between assemblies, enabling direct synaptic strengthening to occur between two assemblies without interference. Oscillating between forward and backwards replay of SWRs can induce bidirectional strengthening, creating strong reciprocal connections between percepts that reflect symmetric relationships.

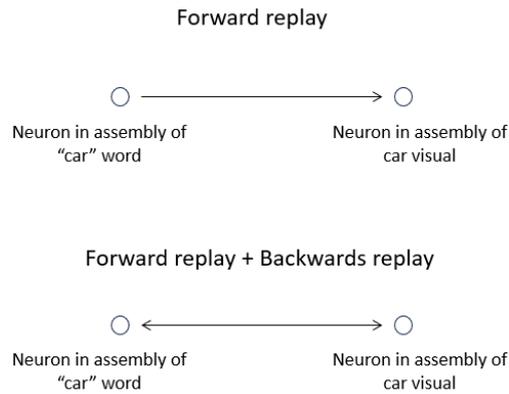

**Figure 3.4:** Strengthening of two assemblies which are conceptually related through forward and backward replay during hippocampal SWRs.

We model cell assemblies (encoding percepts) as graphical nodes, while edges represent strengthened synaptic connections, encoding some form of relationship between percepts. This relation may pertain to any arbitrary real-world relationship: cause-effect, implication, super-ordinality, synonymity, categorical, temporal etc. (**Figure 3.9**).

The reliable activation of one percept by another indicates repeated strengthening of their synaptic connections, suggesting that the two percepts consistently co-occur in the physical world and may share a strong conceptual relationship. The table below identifies specific pairs of percepts that share strong conceptual relations and are likely to exhibit high, consistent real-world co-occurrence.

## Social

| Relation | Percept 1 | Percept 2 | Percept 3 | Concepts represented by a combination of percepts |
|---|---|---|---|---|
| Similar definition | Big | Large | Huge | Concept of being big |
| Modal counterpart | Ice cream (visual) | Ice cream (verbalised word) | Ice cream (written word) | Transmodal concept of ice cream |
| Associative relation | lock | key | | Concept of associative/ functional class |

## Natural

| Relation | Percept 1 | Percept 2 | Percept 3 | Concepts represented by a combination of percepts |
|---|---|---|---|---|
| Similar entity responsible | Appearance of the object | Sound that the object makes | Tactile feel of the object | Concept of the object |
| Thematic | Pen | Books | Teacher | Concept of a classroom |
| Emotional | Anger (Emotional) | Negative reward | Alienation | Concept of an adverse emotional event |

## 3.4 Representation of percepts and percept prototypes

Complex concepts such as <justice> or <credibility> are likely found in a-modal areas, with evidence suggesting that the a-modal area is likely centred on the ventrolateral ATL (Ralph et al., 2017). For complex concepts, integration across modalities is required to fully embody them. <justice (word)> may share thematic relation to <gavel (vis)>, <fairness>, <equality>, <guilt (emo)> or <bias (valence)>.

Modal-specific areas of the ATL are used to represent concepts that are grounded in a particular modality. For example, the concept <a woman dancing to music> cannot be represented in the static visual modality of the ATL, but is more representable through dynamic visual movements. Similarly, the percept of <Nocturne Op. 9 No. 2> is more grounded in audition, or linguistically through its lexical label, with minimal representation in other sensory domains.

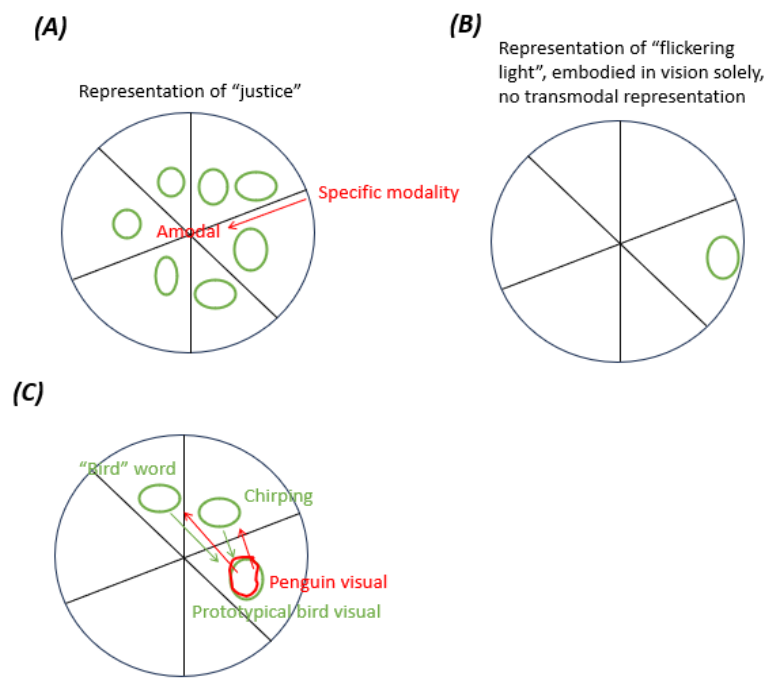

**Figure 3.5: (A)** Abstract a-modal concepts such as <justice> and <creativity> are represented closer to the centre of the graded ATL hub. **(B)** Percepts that are grounded in a specific modality are more likely to be represented by neurons closer to the inputs of that sensory modality. **(C)** The percept of <penguin (vis)> may activate <bird (lex)> and <chirping (aud)>, which may activate the bird visual prototype back in the visual modality.

A few terms are defined. Firstly, the collection of activations of all neurons when being presented with a percept stimulus will be termed the percept prototype. Secondly, the collection of weights for all complementary and singular inputs, from one percept prototype to a constant representation, where the constant representation refers to an idealized percept with no variation or noise throughout its apprehensions, will be termed the relational prototype. Suppose bird $X$ possess visual features $a, b, c$. Bird $Y$ with features of $a, b, d$. Bird $Z$ with features $a, d$. $a$ activates neuron $A$, $b$ activates neuron $B$, $c$ activates neuron $C$, and $d$ activates neuron $D$. The bird visual prototype will thus be, $A = 3, B = 2, C = 1, D = 1$, while the visual relational prototype of birds will be, $AB = 2w, AD = 2w, BD = w, BC = w, A = 3w, B = 2w, C = w, D = 2w, w$ referring to a unit of weight. The creation of this prototype may involve the presentation of <bird (vis)>, followed by a short delay and the presentation of <bird (lex)>, where <bird (lex)> is presented without variation across all trials.

One-for-one hetero-association, a process characteristic of automatic retrieval, activates a transmodal conceptual representation via spreading activation to related percepts. For example, the visual percept of a car may automatically retrieve the sound of its engine, the smell of exhaust, or the lexical label "car". We term this process "relational propagation", defined as the hetero-association of percepts to all related percepts through spreading activation.

However, one-for-one hetero-associations are limited by a lack of contextual constraints. More nuanced and accurate relational propagation is achieved through contextualization—specifically through many-to-one hetero-associations that integrate multiple related percepts. For example, <farm> should not hetero-associate to <chicken> based solely on the broad conceptual category like agriculture, as it is not specific enough. Instead, this association may be established only in a context where the individual is prompted with the query <what belongs on a farm?>, or paired with supplementary percepts such as <animal> or <livestock>. Contextualization through the provision of supplementary percepts is an inherently memory-based process. In the following section, relational propagation through the contextualization provided by supplementary percepts (many-to-one hetero-associations) will be explored.

## 3.5 Contextualized hetero-associations (many-for-one)

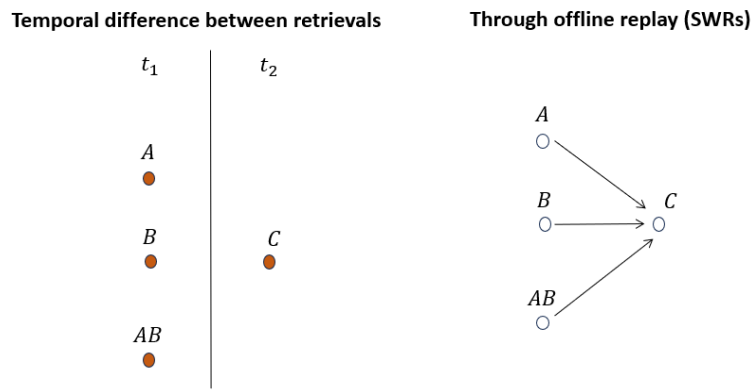

**Figure 3.6:** Hippocampal sequences encode the order of retrieval of percepts. During offline replay, the compressed sequential firing of neuronal assemblies encoding these percepts induces synaptic strengthening between them, relating the percepts together. In the illustration above, A and B are input assemblies and C is the target assembly, encoding input percepts and target percepts, respectively.

The co-activation of two assemblies encoding percepts that occurred seconds apart requires the temporary storage and maintenance of percepts, whether sustained by persistent activity in prefrontal recurrent networks (Constantinidis et al., 2018) or by activity-silent short-term plasticity mechanisms (Kozachkov et al., 2022).

Unlike the prefrontal cortex (PFC), the ATL is not known for maintenance of working memory. While attractor states holding information can arise from local recurrent structures, recurrence in the ATL is better suited for pattern completion and semantic priming than for holding concrete representations (Munakata et al., 2011). For prior percepts to be accessible to the ATL, the PFC supplements prior percepts stored in working memory.

Because the PFC's primary function is to coordinate goal-directed actions, percepts are contextually pruned and competitively gated in working memory (Chatham et al., 2014) to ensure task relevance. In controlled semantic cognition (CSC), the PFC acts as an executive controller, providing task-relevant signals to the ATL to bias retrieval of conceptual representations (Ralph et al., 2017). We suggest that this retrieval signal is analogous to supplementary percepts, which may include both direct activations and activation biases for percepts in the ATL. Since PFC initiated retrieval is goal-directed, it may be interpreted as controlled retrieval.

Similar to single neurons, the presynaptic assembly will be termed the input assembly, and the postsynaptic assembly will be termed the target assembly, encoding input percepts and target percepts, respectively. For the case of assemblies, CPH predicts that the activation of neurons within the target assembly shortly after the activation of neurons within input assemblies will cause neurons in the input assemblies to experience plasticity such that subsequent firing of any subset of neurons within the union of input assemblies will induce an increased excitatory signal to the target assembly. Complementary inputs can thus be created between input assemblies during their co-activation, constructing complex input-percept to target-percept relations. The complementary input will be labelled as a concatenation of both input assemblies, as shown in **Figure 3.6.**

Complementary inputs for assemblies provide functional versatility for modelling real-world conceptual relations. For example, the target <ice cream> is related to the inputs of <cone>, <chocolate>, <cold>, and <treat>. The weight between <cone> and <ice cream> is likely to be small, as a cone is used in many different contexts, such as geometry, traffic and sports. Likewise, the weight between <cold>, <chocolate>, <treat>, and <ice cream> is also small, as each input can be used to describe many things.

We identify two percept combinations with high complementary weights to the target: <chocolate> and <cone>, and <cold> and <creamy>. The inputs <chocolate> and <cone> generate the concept of a cone-based dessert, while <cold> and <creamy> describe the sensory properties of ice cream, having high complementary weight to the target. Input complements enable combinations of percepts to produce a supra-linear response when conceived together, exceeding the sum of their individual activations.

## 3.6 Supervised learning

For two conceptually related percepts to be encoded in hippocampal traces and replayed to the cortex, they must occur in a precise sequence within a tightly constrained temporal window. However, this condition is rarely met in real-world situations. For example, <apple (vis)> does not reliably precede <apple (lex)> in natural experience, as one does not encounter the apple word every time one observes the image of an apple.

This limitation can be circumvented through supervised learning, where an instructor deliberately presents correlated percepts in close temporal proximity, such that the learner can associate the percepts together. The simultaneous presentation of the image of an apple and its verbal label provides the necessary conditions for effective hippocampal encoding and memory consolidation.

When advancing beyond supervised learning into noisy real-world environments, previously learned relations can still be maintained through self-reinforcement. This process constitutes a positive feedback loop in which strongly connected percepts exhibit a higher probability of mutual activation, further strengthening their connections when activated. Balanced by adequate homeostatic plasticity and synaptic scaling, the feedback loop facilitates the long-term retention of percept relations post-consolidation. Consequently, unless novel information fundamentally contradicts the established relation, SWR involvement is no longer necessary after consolidation (MacDonald et al., 2011).

## 3.7 Automatic and controlled semantic retrieval

To reinstate, controlled retrieval is characterised by effortful, contextualised recall mediated by working memory and conducted through many-for-one relational propagation. In contrast, automatic retrieval is a bottom-up, perception-driven process characterised by one-for-one relational propagation. The former supports goal-directed problem solving while the latter supports object recognition, semantic conceptualisation and automatic allusions to closely related percepts through relational propagation of a single percept. Crucially, because both forms of retrieval leverage the same learned real-world relations between percepts, they can be multiplexed within the same neuronal population in the ATL.

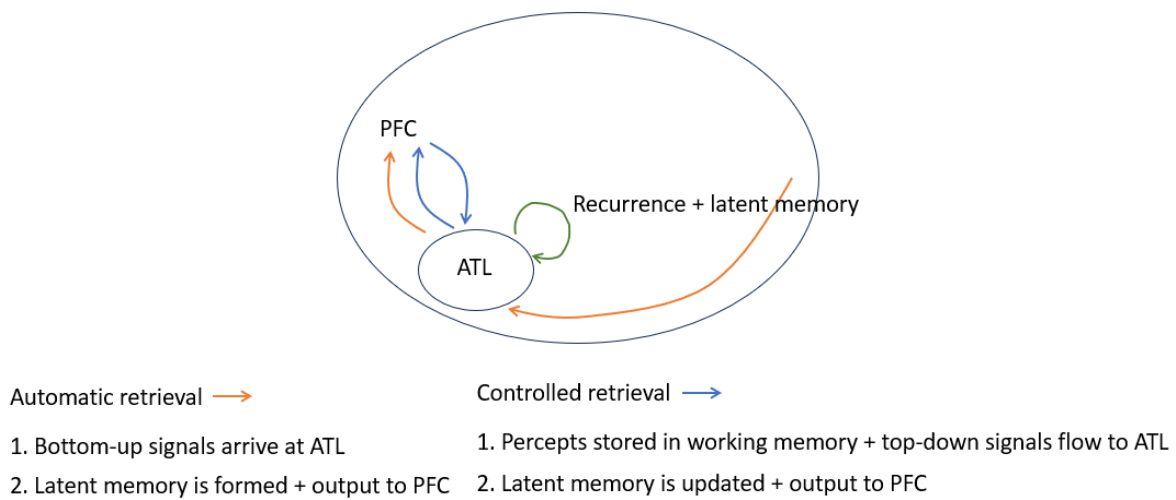

**Figure 3.7:** An illustration of the distinction between controlled and automatic retrieval. Automatic retrieval is driven by bottom-up stimulation, whereas controlled retrieval is initiated by top-down signalling from the PFC. Recurrent connectivity within the ATL enables the short-term sustenance of percepts for effective relation propagation.

## 3.8 General neural properties

Connections between neurons convey universal relations. While relationships between concepts are often described by logical specifications such as <A implies B>, <A causes B>, and <A temporally precedes B>, these logical specifications will not be conveyed in real-world situations, a key flaw of utilizing propositional semantic graphs in early cognitive architectures such as SOAR and ACT-R.

The relational specificity offered by logical specifications can be compensated for by simply providing sufficient supplementary percepts to ascertain the relation. For example, consider two relations, <seed leads to growth> and <seed implies life>. Introducing the supplementary percepts of <water> and <germination> will allow <seed> to activate <growth> and not activate <life>, while introducing the supplementary percepts of <animals> and <living things> will allow <seed> to activate <life> and not <growth>. Supplementary percepts resolved the ambiguity of relations, without requiring logical specifications.

Similar to relations, the identity of a percept can be clarified by supplementary percepts as well. For instance, <bow (lex)> could mean the act of bowing or weaponry, clarified by whether the supplementary percepts of <performance> or <hunter> activate.

Lastly, neurons are powerful generalizers. Observing any new percept connects its representation to a larger preexisting network, enabling its characteristics to be understood immediately while further expanding the existing network. In theory, any percept will only have to be observed once for it to be understood through consolidation into the existing network.

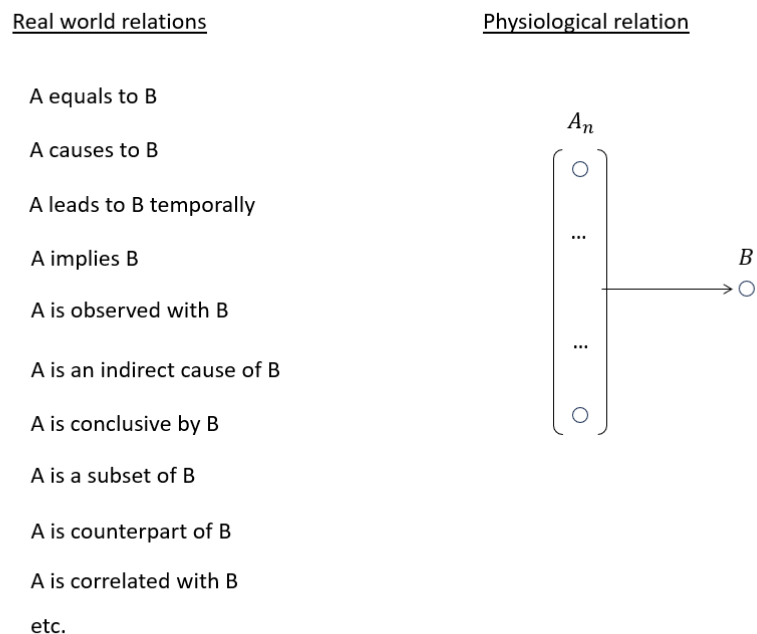

**Figure 3.9:** Strengthened connections between two neurons convey universal real-world relations

## 3.9 Representation of complex percepts

While the ATL can represent simple percepts such as <car> or <dog>, it can also represent complex, multi-component concepts like <a man painting an artwork> or abstract ideas such as <expression of artistic identity>. Complex percepts such as <imagination> or <justice> may be object-independent, lacking any direct physical instantiation.

We propose that representation of complex percepts in the ATL occurs through three predominant mechanisms: (1) grounding them in appropriate sensorimotor modalities, (2) repeated exposure to the complex percept in its exact form, and (3) through percept combinatorics.

The first mechanism involves representing the complex concept in its appropriate modality. The complex percept of <a beautiful night sky> can be more effectively represented in static visual modality than in auditory, linguistic, and somatosensory modalities. Similarly, the complex percept of <a man painting an artwork> can be represented by integrating a static visual form of a man sitting behind a canvas, with the dynamic visual motion of the painting action.

The second mechanism allows complex percepts with physical instantiations to be represented through repeated exposure to their consistent, exact physical form. For example, <a man painting an artwork> can be represented in static visual modality through repeated observation of a scene depicting a man sitting in front of a canvas holding a paintbrush. This repeated exposure facilitates the development of specialized neuronal tunings that encode features unique to the composite percept itself – features that do not exist within the constituent objects in isolation. Consequently, the representation of <a man painting an artwork> will encompass three neuron populations, one population representing <man>, one population representing <canvas>, and another population representing the features unique to <man painting an artwork>. Neurons tuned to features unique to <man painting an artwork> develop a conjunctive code of both constituents. Conjunctive codes for complex percepts are critical to establish distinction from their constituents, as constituents can have different meanings as compared to their entirety.

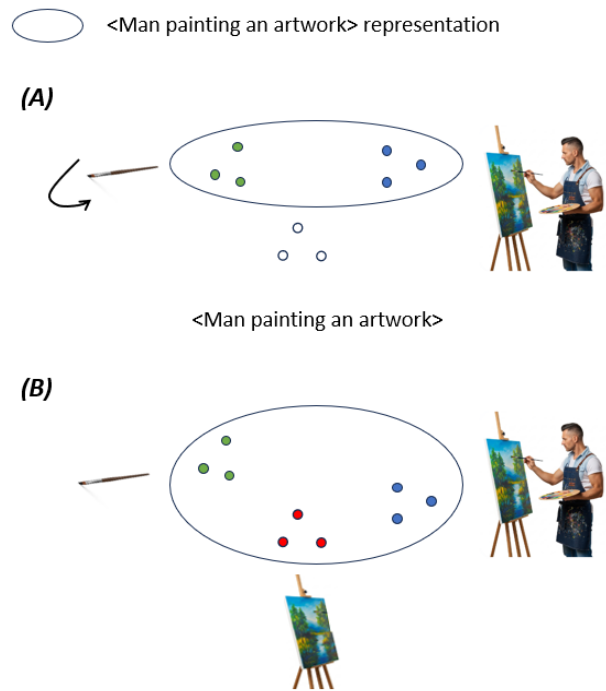

**Figure 3.10: (A)** The grounding of the complex percept <man painting an artwork> in its relevant modalities, namely through visual motion stimulus of the action of brushing, and static visual stimulus of a man painting a canvas, rather than the lexical label of <man painting an artwork>, which is less frequently encountered. **(B)** The representation of <man painting an artwork> is created through repeated exposure to the visual stimulus of a man painting on a canvas. The representation is composed of three assemblies: one for <paintbrush>, one for <canvas>, and a third that encodes the features of the unified scene <man painting an artwork>.

The third mechanism of representing complex percepts involves the combinatorial integration of less complex percepts. For instance, <a man painting an artwork> can, in principle, be represented as a combination of <man> and <canvas>. However, this approach is problematic, as it relies on the unrealistic assumption that this specific combination can only yield one interpretation. This leads to ambiguous and contradictory associations; for example, the same combination may be activated by both the peaceful act of painting and the violent act of slashing a canvas, resulting in conflicting associations of <violence> and <peaceful>.

A more effective and disambiguating strategy is to combine more specific percepts such as <man holding a paintbrush> and <canvas with paint>. This refined combination decorrelates the target representation from irrelevant stimulus, as the image of a man slashing a canvas will no longer activate these constituent percepts.

These constituent percepts must be represented either through direct, repetitive exposure (Second method) or through other percept combinations (Third method). Ultimately, all percepts will still be represented through the creation of percept-specific tunings (Second method). We propose that, in general, the more complex the percept, the greater the number of percepts combined and the larger the neuron population involved in representing it.

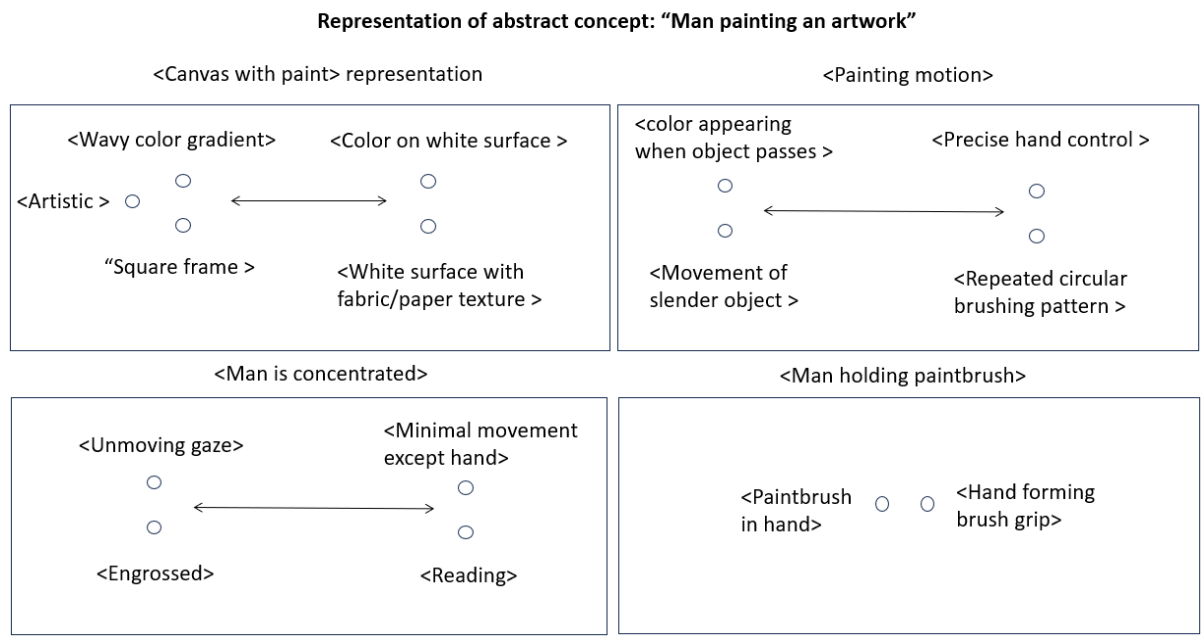

**Figure 3.11:** Complex percept of <man painting an artwork> being represented by combinations of percepts, where certain combinations can be substituted for other combinations. Reciprocal connections indicated by double-headed arrows suggest a strong relationship between percept groups (combinations of percepts), where one percept group may activate the other. The concept is ultimately represented by a totality of all percepts in the diagram.

The third mechanism can also leverage consolidated relations between constituent percepts and generalize across multiple percepts. Relational propagation between percepts allows for specific percepts to be activated in a wider range of situations, without requiring a specific stimulus, which may be absent due to noise.

The capacity for cross-contextual integration, in which percepts are activated across diverse situations, is similar to the principles of invariance explored in the VPS. However, while invariances in the VPS are well-defined —<positional>, <viewpoint>, etc. —invariances for conceptual percepts are often challenging to articulate linguistically, as they lack a common defining term. For instance, as referenced in **Figure 3.11**, <canvas with paint> representation may be invariant across <expression enabled by the material for arts>, while <man is concentrated> representation may be invariant across <studious pupil who pays close attention>. Critically, these conceptually based invariances emerge naturally within the ATL as an inherent property of forming relational networks between percepts through experience.

To achieve effective relational propagation, the PFC mediates the representations present in the ATL at a given point in time, preventing interference between potentially conflicting percepts. Additionally, it regulates attention to external stimuli, controlling the percepts perceived and represented. For instance, the percepts of <square frame> and <engrossed> may relate to irrelevant percepts of <appraising a painting> or <admiring an artwork>. Interference can be mitigated by selectively attending to different aspects of the image at separate moments, such as attending to the canvas first, retrieving <square frame> and all related percepts, before attending to the facial features of the man, retrieving <engrossed> and all related percepts. Temporally staggering the retrieval of two representations in ATL allows them to be bound coherently while minimizing interference from irrelevant, disparate representations.

## 3.10 Prevention of catastrophic interference

The complementary learning systems (CLS) framework that describes the hippocampus as a fast-learning system that rapidly encodes recent episodic memories, while the cortex is a slow, gradual learner that extracts statistical regularities from episodic memories to form knowledge schemas (O'Reilly et al., 2014; Squire et al., 2015) – structured frameworks representing knowledge to influence behavior.

CLS emphasizes the prevention of catastrophic interference, a phenomenon where learning may overwrite existing schemas. First, two separate learning systems accommodate both fast learning of one-time events and slow updating of general knowledge (Quiroga, 2020). Second, the hippocampus's ability to replay episodic memories offline during SWRs enables the interleaving of new memories with previously acquired episodic memories or other related consolidated schemas (O'Reilly et al., 2014; McClelland et al., 1995). Third, pattern separation, achieved through sparse activity in the dentate gyrus, enables the storage of multiple memory traces with minimal interference (Johnston et al., 2016).

While these mechanisms may be incorporated into our model, we suggest reasons why catastrophic interference may not occur even in their absence. First, while two assemblies may overlap due to a large number of shared features, their representations are orthogonalized by the nature of complementary inputs.

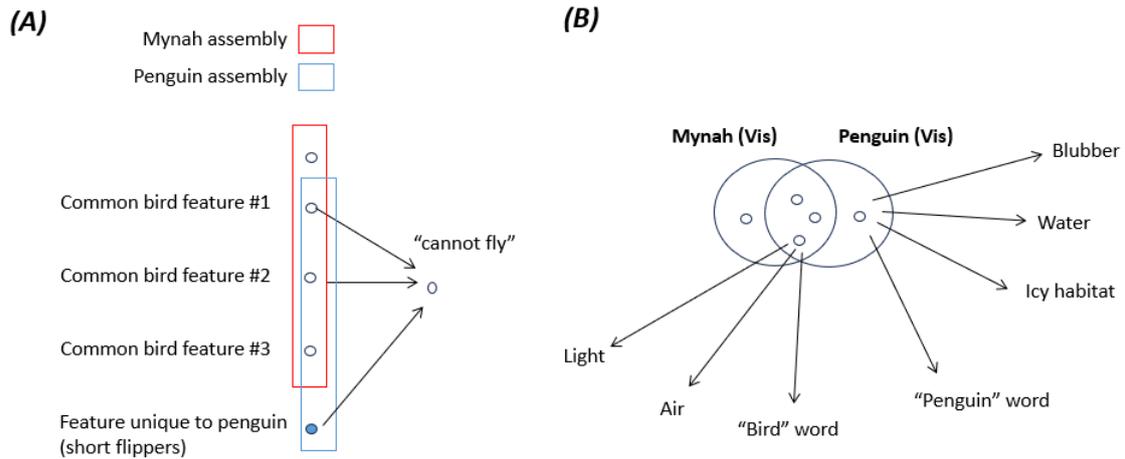

**Figure 3.12: (A)** Assuming that there are three overlapping neurons between the penguin and mynah assembly, one neuron unique to mynahs, and one neuron unique to penguins, perhaps encoding the feature of short flippers. **(B)** The bird prototype may activate the percepts <bird (word)>, <air>, and <light>. Neurons unique to each bird type may activate bird-specific relational targets.

Penguins and mynahs are both birds, and they are visually similar, relative to some arbitrary object. Suppose three out of every four neurons are shared between their representations, implying a three-quarters overlap as shown in **Figure 3.12A**. During the strengthening of the assembly to the <cannot fly> assembly, a total of sixteen complementary inputs are activated for every subset of inputs. Assuming that each of the sixteen complementary inputs has equal weight to the target. Suppose any one of the neurons within the penguin assembly does not activate. In that case, eight of the sixteen complementary inputs will not activate, halving the required presynaptic excitation required to activate the target, assuming no redundancy in complementary input excitation and ignoring single input excitation. The significant decline in input excitatory signal for <cannot fly> assembly without the activation of the penguin-specific neuron, orthogonalizes the representations of mynah and penguin despite a high degree of overlap. Additionally, the presence of multiple penguin-specific assemblies and multi-synaptic connections between neurons within the input and target assemblies can further increase the need for a specific assembly's activation, thereby preventing catastrophic interference.

Secondly, while two representations overlap in one modality, the same may not be true across other modalities. While penguins and mynahs may be visually similar, they differ in their linguistic representation, i.e. <penguin (lex)> vs <mynah (lex)>. Thus, the transmodal representations of penguins and mynahs are naturally orthogonalized, unlike their unimodal representations.

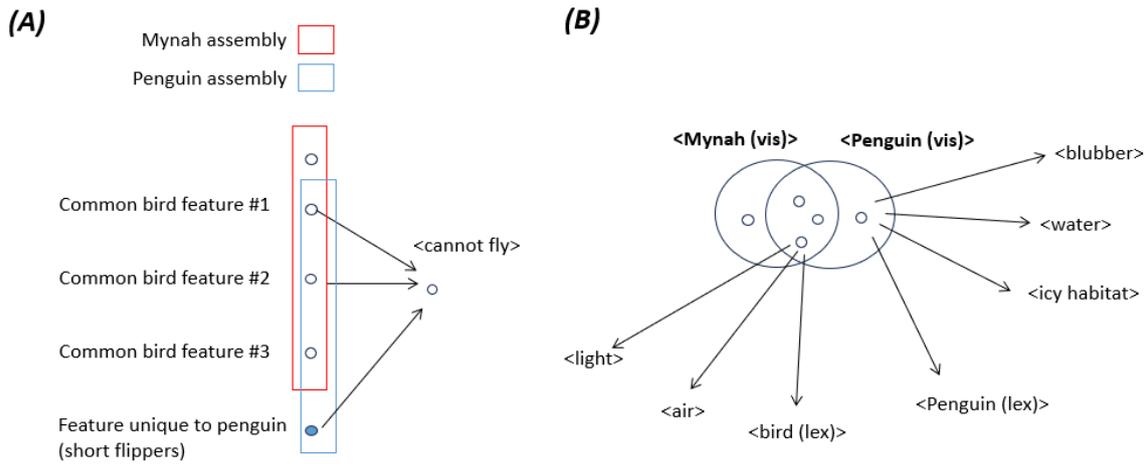

**Figure 3.13:** Dotted arrows indicate weak connections, solid arrows indicate strong connections, and blue circles represent neurons unique to a particular bird. **(A)** A simplified illustration of the <bird (vis)> relational prototype, represented by the connection weights to <bird (lex)>. The unique features of a bird and its complementary inputs have weak connections, depicted by dotted arrows, as compared to common features across most birds that have strong connections. **(B)** Common features of a percept are strongly strengthened to common features of other percepts, while less common features are weakly strengthened to other less common features. The unimodal percept prototypes of <bird (aud)>, <bird (vis)>, and <bird (lex)> collectively form the transmodal percept prototype of <bird>.

Nonetheless, orthogonalization does not imply a lack of generalizability. Firstly, percepts are orthogonalized insofar as their unique representations are; overlapping representations may still be used for relational propagation. Relations encoded by connections of an overlapping representation are generalized across all percepts containing that overlapping representation. Secondly, generalizations are strictly veridical, where every experience contributes a unique cortical trace, which is then summed and averaged into a generalized relational map between percepts. Relations conveyed by overlapping traces are not restructured, but integrated across experiences.

Previously, we discussed that multiple variations of <bird (vis)> can create a bird relational prototype, represented by the weights of <bird (vis)> to some related percept, namely <bird (lex)>. Critically, it was assumed that <bird (lex)> has no variation across trials. In reality, <bird (lex)> will possess variations across idiosyncratic apprehensions, through differences in handwriting or verbalization of the word. More generally, the complementary weight of A, B to every neuron in C, where A, B and C are arbitrary sets of neurons, is correlated to the probability of the activation of C, given the activation of A and B.

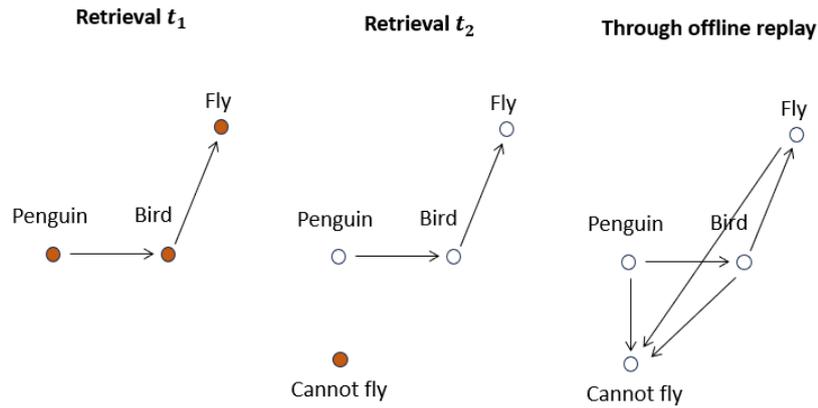

**Figure 3.14:** An illustration of steps leading to the strengthening of <penguin> to <cannot fly>, upon learning that penguins cannot fly, assuming that the individual knows penguins are birds. **(Left)** Initially, <penguin> is strongly connected to <bird>, which is strongly connected to <fly>. The activation of <penguin> at $t_1$ will activate <bird> and <fly>. **(Centre) <**Cannot fly> activates after at $t_2$. **(Right)** Through offline replay, <bird>, <penguin> and <penguin> are strengthened to the <cannot fly>.

Consider a consolidated schema in which all birds are assumed to fly. When a penguin is introduced as a counterexample, the consolidated schema is violated. To prevent catastrophic interference, it must be ensured that <cannot fly> does not activate in the case where <mynah> activates, and <fly> does not activate when <penguin> activates. The former is illustrated under **Figure 3.15**, the latter in **Figure 3.16**.

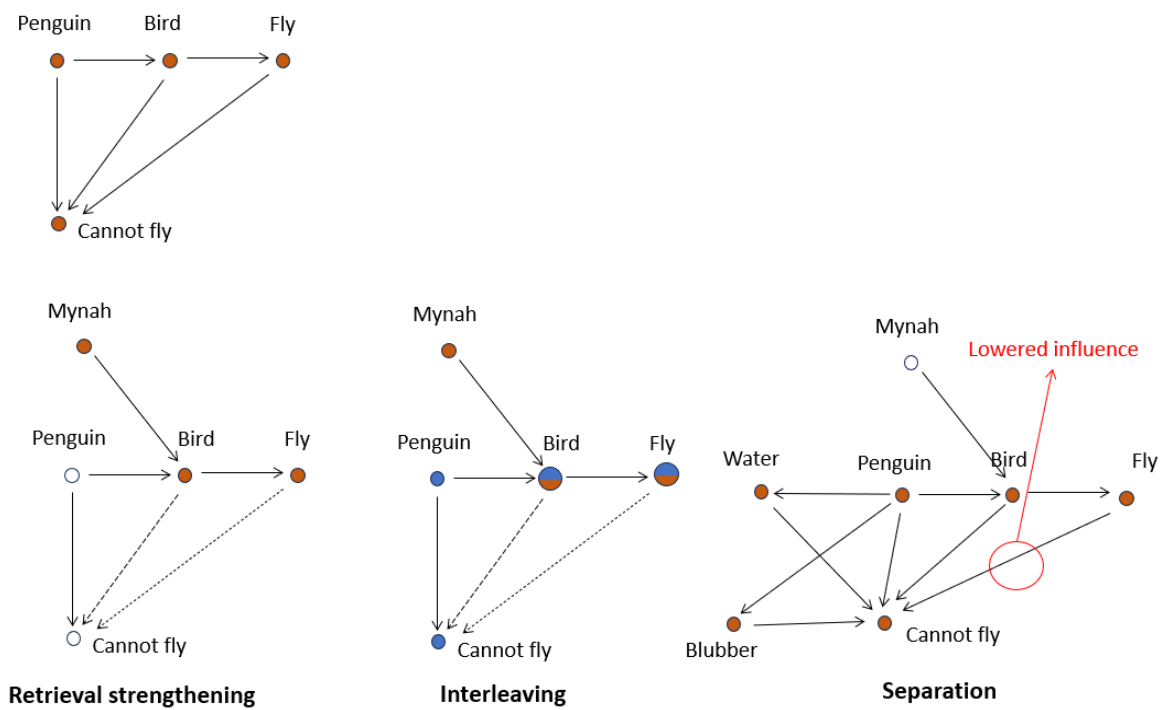

**Figure 3.15:** Delineation of three possible mechanisms to prevent <cannot fly> from activating when <fly> activates. Dotted lines indicate weakening of connections. **(Left)** <mynah> can activate <bird> and <fly>, <cannot fly> will not activate without <penguin>. Following

repeated retrievals or through offline replay, consolidating <bird> to <fly>, the connections between <bird>, <fly>, and <cannot fly> are weakened, while the connections from <penguin> to <cannot fly> remain strong. **(Centre)** During offline replay, the selective strengthening of <penguin>, <bird> to <cannot fly> is interleaved with the selective strengthening of <bird> to <fly>. As a result, the connections between <bird>, <fly>, and <cannot fly> will be weakened due to uncorrelated activation, while the connection between <penguin> and <cannot fly> will be strengthened. **(Right)** Relating more percepts unique to <penguin> decreases the excitation influence of <fly> to <cannot fly>. For instance, <penguin> relates to <water> and <blubber>, which <bird> and <fly> are not related to. When <penguin> activates, it activates <fly>, <blubber>, <bird>, <water>, which are strengthened to <cannot fly>. As <fly> only accounts for one in five strengthened percepts to <cannot fly>, their activation has less influence over whether <cannot fly> activates, decorrelating <fly> from <cannot fly>.

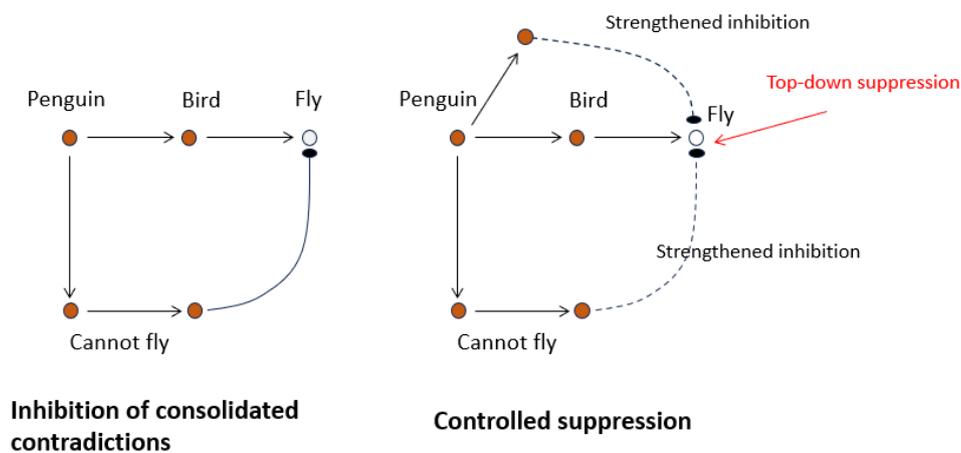

**Figure 3.16:** An illustration of two mechanisms to prevent <fly> from activating during the activation of <penguin>. **(Left)** Consider the existence of an inhibitory neuron that strongly suppresses the percept of <fly> when <cannot fly> activates, <fly> will not activate when <cannot fly> activates, decorrelating their activations. As a result, when <penguin> activates, <cannot fly> activates, suppressing <fly> from activating. **(Right)** If direct neural inhibition is absent between <fly> and <cannot fly>, top-down control suppresses <fly> during the retrieval of <penguin>. When inhibitory neurons with inputs <penguin> and <cannot fly> are activated, it prevents the activation of <fly>. Through inhibitory STDP, inhibition towards the postsynaptic <fly> is strengthened.

Leveraging existing schemas when encountering novel information enables faster integration than when no relevant schema is available. For instance, learning that penguins cannot fly can occur much faster if the penguin schema already exists, thereby enabling the strengthening between <penguin> and <cannot fly> instead of forming an entirely new schema for penguins. Over time, the accumulation of experiences may introduce more percept unique details, orthogonalizing representations and leading to accelerated schema integration. Indeed, this is often supported by research showing a faster rate of consolidation for novel information when a relevant schema has been previously created (Tse et al., 2007).

# 4. Module 3: Prediction module

The predictor module encompasses two submodules. Submodule 3.1 is used for generating generalized predictions, while Submodule 3.2 is used for generating contextualized episodic predictions.

## 4.1 Module 3.1: Prediction module (generalized temporal sequences)

Corresponding anatomical region(s): Distributed brain regions containing time and sequence cells, ATL

The predictor module generates a prediction of the future state given the current state. The current state refers to a temporally ordered sequence of percepts that has been observed, while the future state corresponds to a percept representing the temporal successor of the observed sequence.

In the physical world, stimuli arrive at different times in the sensory cortices, where they are processed and represented. Time cells activate at different timings within a sequence, timestamping incoming percepts and encoding a conjunction of elapsed time and percept identity (Tiganj et al., 2017). Sequence cells selectively activate in response to percepts arriving in a defined order, encoding ordered sequences.

The conjunctive representation of percept identity and sequence will be designated as a temporal percept. A temporal percept will be denoted as <apple, 1> for time cells, where the second argument indicates elapsed time, <apple, banana> for sequence cells, indicating ordered percept pairs. Temporal sequences will be denoted as <<banana, 1>, <apple, 3>: <lemon>>, with the first and second elements being temporal percepts, and the third being a regular percept.

Activation of temporal percepts within a temporal sequence can lead to the activation of a regular percept. We can also conceptualize a temporal sequence as a regular percept. Consequently, any regular percept activated by the temporal sequence has some form of real-world relation to it – for instance, representing the subsequent element of the sequence, or a label of the meaning or type of sequence.

Similar to module 2, generalized and decontextualized temporal sequences are consolidated into the cortex through SWRs, shown in **Figure 4.1**. We further propose that neither temporal percepts nor percepts can hetero-associate to a temporal percept, with a lack of evidence to support the capacity of cortical time cells to directly induce the firing of other time cells.

This submodule consists of two systems, the temporal percept space and the percept space. The temporal percept space encompasses distributed brain regions containing time and sequence cells, thereby capable of supporting temporal percepts. The percept space encompasses regions such as the ATL, sensory cortices, emotional networks, and the motor cortex, among other areas representing percepts. Percepts from the ATL will be termed as ATL percepts, percepts encoding sensory stimuli as sensory percepts, and so forth.

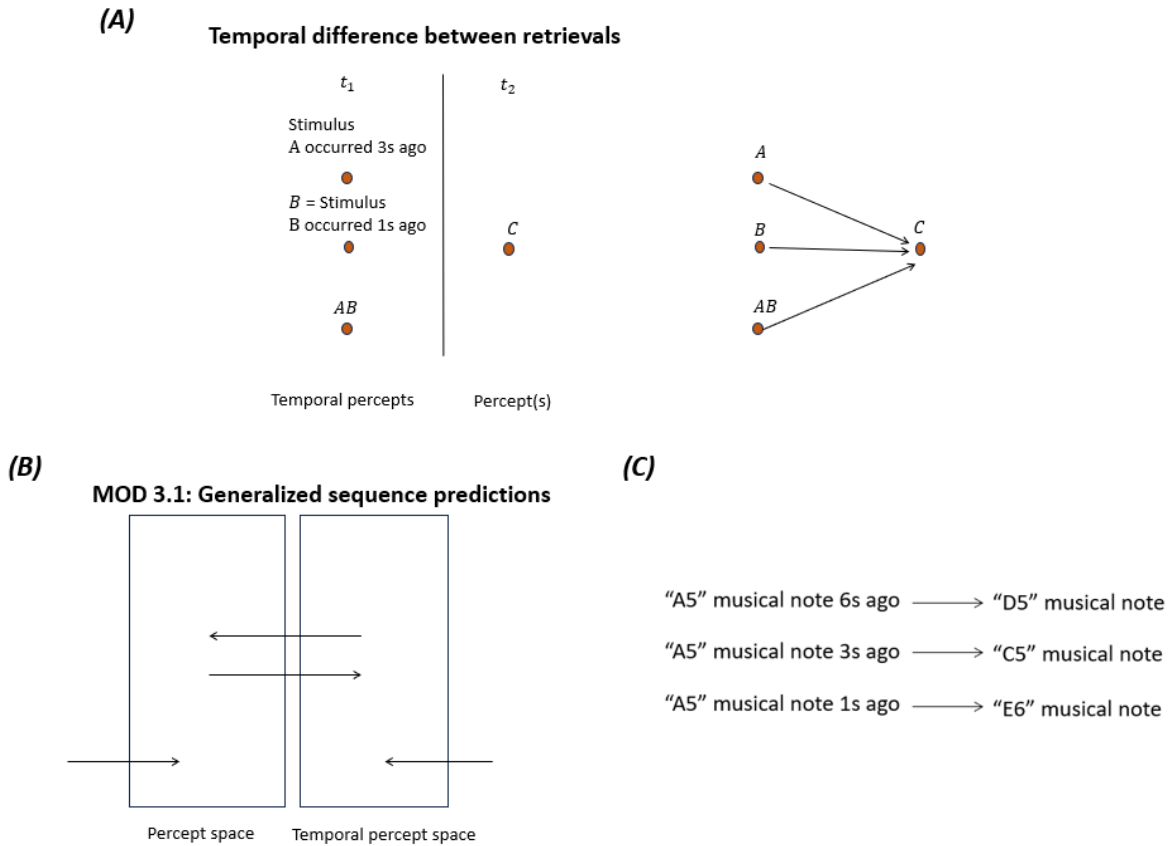

**Figure 4.1: (A)** The strengthening between two percept contexts occurs through offline replay. Consider time cell A, which activates when stimulus A occurs three seconds prior, and time cell B, which activates when stimulus B occurs one second prior. Consequently, the activation of temporal percepts activates C. Through SWRs, A and B, coupled with the complementary input AB, will be strengthened to C. **(B)** Schematic diagram of module 3.1, consisting of the temporal percept space and the percept space. Arrows indicate projections between both regions that subserve relational propagation. **(C)** An example of how precise timings of the same stimulus are important in auditory sequences, where the same stimulus can have different implications depending on how recently it was perceived.

Similar to VPS neurons, time cells also possess temporal tuning curves, providing tolerance for the exact timing of stimulus occurrence. With sufficient time cells, their tuning curves can span temporal intervals, enabling the extraction of complex temporal rules through different combinations of firing patterns. Exact timing or order of temporal percept activation is important, as percepts occurring at different timings or in different orders can activate different targets. One example could be music, auditory sequences that are highly sensitive to temporal order, as shown in **Figure 4.1B**.

### 4.1.1 Complementary inputs for temporal percepts

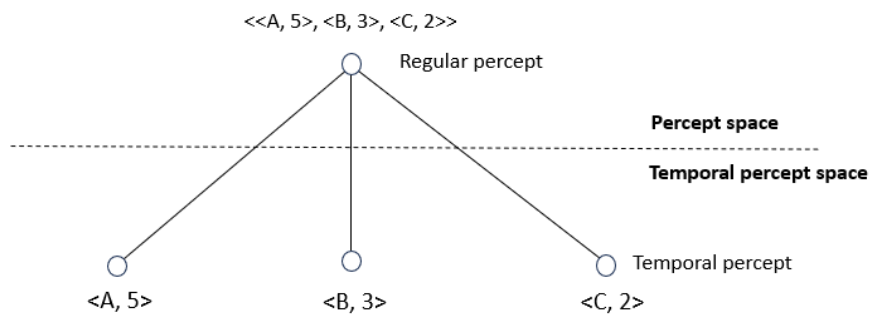

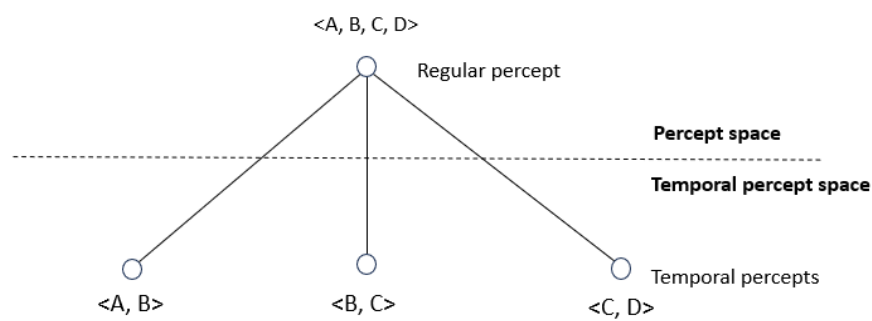

**Figure 4.2: (A)** An illustration of a relation between three temporal percepts in the temporal percept space and a target percept in the percept space. Temporal percepts can form complementary inputs, and different complementary inputs may possess different weights to the target percept. **(B)** <A, B> represents the ordered sequence of percept A occurring before percept B. Here, temporal percepts are encoded by sequence cells.

Each temporal sequence also has a prototype, characterized by the weights of all temporal percepts and percept complements to the target percept for every variation of the temporal sequence observed before the target percept. The complementary inputs with the strongest weights represent the most frequently occurring combinations. The probability of activation of the target percept correlates with the probability of the temporal sequence's activation and the probability of its constituent temporal percept activations.

We propose that sequence cells can encode temporal percepts of more than two elements. Overlapping elements within temporal percepts can give rise to complex relations. For instance, a sequence cell may encode a temporal percept of <A, B, C>, and another may encode <A, B>, each possessing different weights to the target. Although elements overlap, such redundancy enables extensive combinatorial variability through which temporal rules can be expressed. The complementary effect of both time and sequence cells can further diversify relational structures capable of encoding abstract temporal rules.

Relational propagation can include both temporal percepts and regular percepts. Combinations of temporal percepts and percepts may form complementary inputs to activate other regular percepts. Consider a scenario in which pressing a button while a bell is ringing causes a box to appear 4 seconds later. The activation of <box> when the temporal percept

<pressing button, 4>, contingent upon the activation of <bell ringing>. The complementary input between <pressing button, 4> and <bell ringing> encodes the complementary relationship between them.

In summary, the relational propagation between the temporal percept space and percept enables the encoding of complex temporal rules. The following section elaborates on this framework using real-world examples.

## 4.1.2 Generalizability of temporal sequences

Temporal sequence in the real world can involve complex sequences of events:

1. Dog growls → Barks → Lunges → Bites and chases: Aggressive dog (Descriptor)
2. Red light → Cars stop → Green light → Cars move → Yellow light → Cars slow down : Cars stop (Successor)
3. Dark clouds → Thunder → Raindrops → Puddles forming → Clearing skies → Sunlight: Weather cycle (Descriptor)
4. Being seated → Looking at a menu → Ordering food → Eating: Profits, Cutlery, Service (Thematic relation)

Temporal sequences are more powerful than simple prediction generators. Percepts and percept relations consolidated through past experiences can be used to generalize and process novel future experiences through relational propagation.

Suppose an individual had an experience where their pet dog was behaving energetically and scratched a pillow afterward. We assume that the apprehension of these percepts is chronological, namely, <dog> followed by <energetic> followed by <scratching a pillow>. Additionally, <dog> was apprehended 5 seconds before <energetic>, which creates the temporal sequence of <<dog, 5>, <energetic, 0>: <scratching a pillow>> for time cells, and <<dog, scratching pillow>: <scratching a pillow> for sequence cells. After the experience, the temporal sequence is consolidated into the temporal percept space.

The individual also has a pet cat. On a particular day, the cat was observed to be jumping around actively, activating the temporal percepts of <cat, 5> and <jumping, 0>, as <jumping> was apprehended 5 seconds after <cat>. Thereafter, the individual concludes that the <cat may tear the curtain>.

We propose a method for how the individual may arrive at the conclusion, despite not possessing either a consolidated temporal sequence of <<cat, 5>, <jumping, 0>: <tearing a curtain>> for time cells, or <<cat, jumping>: tearing a curtain> for sequence cells. We observe an overarching theme across the two sequences of events: an animal is observed causing damage to furniture.

<dog> can be related, albeit weakly, to <cat> because both belong to the super-ordinal category of household pets. Additionally, other supplementary contextual percepts during controlled retrieval can bias relational propagation to <dog>. For instance, both events occurred in a house, both the cat and the dog are the individual's pets, both of the animals are similar in size, and they are both affectionate, playful, etc. As such, when combined with the supplementary percepts of <my pets>, <affectionate>, <house>, and <playful>, <dog> can easily propagate to <cat>.

Next, <jumping> may associate to the <energetic> percept, as both are movement-related. Activation of related percepts can assist in activating relational propagation targets; for example, <jumping> may activate the related percepts of <movement>, <intense cat movement>, < vigor>, or <fast>. The related percepts, coupled with the supplementary percepts <jumping>, may enable activation of <energetic>.

The actual experience of <cat>, followed by <jumping>, has been transformed into <dog>, followed by <energetic>. As <cat> and <jumping> were perceived 5 seconds apart, <dog, 5> and <energetic, 0> will be activated, assuming no additional delay during relational propagation. As both temporal percepts of <<dog, 5>, <energetic, 0>: <scratching a pillow>> have been activated, <scratching a pillow> will activate.

Finally, <scratching a pillow> may activate the <destruction to furniture>. Coupled with the supplementary percept of <curtain (vis)> derived by attending to the curtain of the house, <destruction to furniture>, <curtain (vis)>, and <cat> can activate <cat may tear the curtain>.

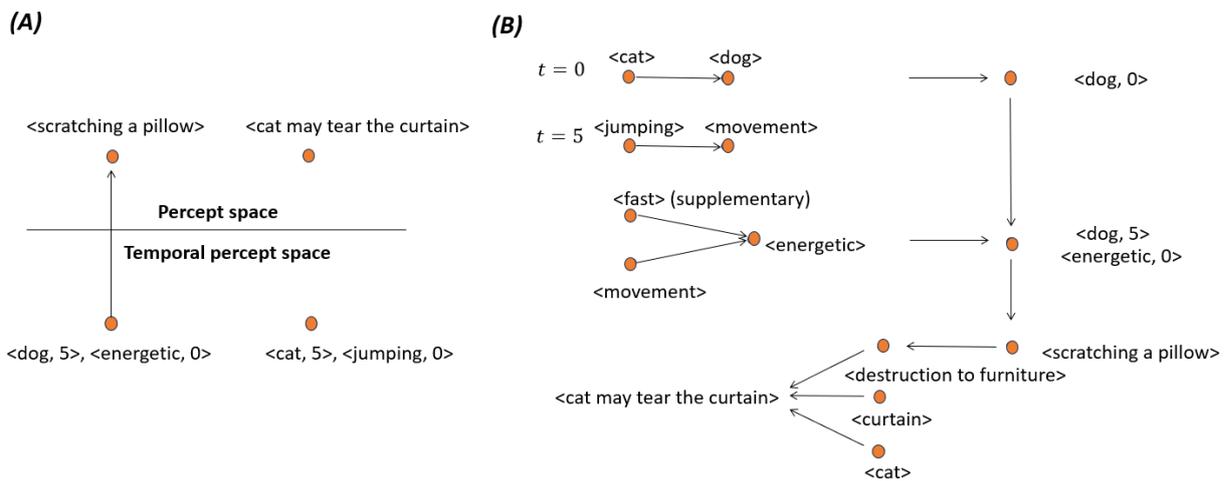

**Figure 4.4: (A)** This illustration demonstrates the goal of activating the target <tearing the curtain>, upon the activation of the temporal percepts of <cat, 5> and <jumping, 0>, given the existence of the consolidated temporal sequence indicated by the solid arrow. **(B)** An illustration of relational propagation steps for the generalization of consolidated temporal sequences. Explanation provided in the main text.

The relational propagation steps identified in this example serve as a reference for how relational propagation may commence. The actual environment may contain numerous supplementary percepts, which may inhibit or amplify the relational propagation targets. As such, the specific targets identified in this example may not activate, depending on the context and environment of the actual experience.

Real-world environments or situations with non-zero likelihood of occurring or encountering serve as a contextual binder for objects or percepts found in or related to them, correlating otherwise arbitrary objects. For example, a library usually contains books, desks, and chairs. If a place has books, there is a higher-than-random chance that it also has desks and chairs, and that it is a library. Likewise, an argument usually entails anger, conflict, and multiple stakeholders. If anger is present in a situation, then there is a higher than random chance that frustration, multiple stakeholders are also present, and the situation is an argument.

This means that within a situation or environment delineated by active pets causing damage to furniture inside a house, provided that this situation is veridical and has a non-zero chance of occurring (pets definitely do cause damage to furniture sometimes), other supplementary percepts in this environment are more likely to amplify, rather than inhibit, relational propagation targets. As such, a greater quantity of supplementary percepts within the environment <cat jumping inside house> increases the probability of deriving similar relational propagation targets, reaching a similar conclusion <scratching a pillow>, despite real-world variability and noise.

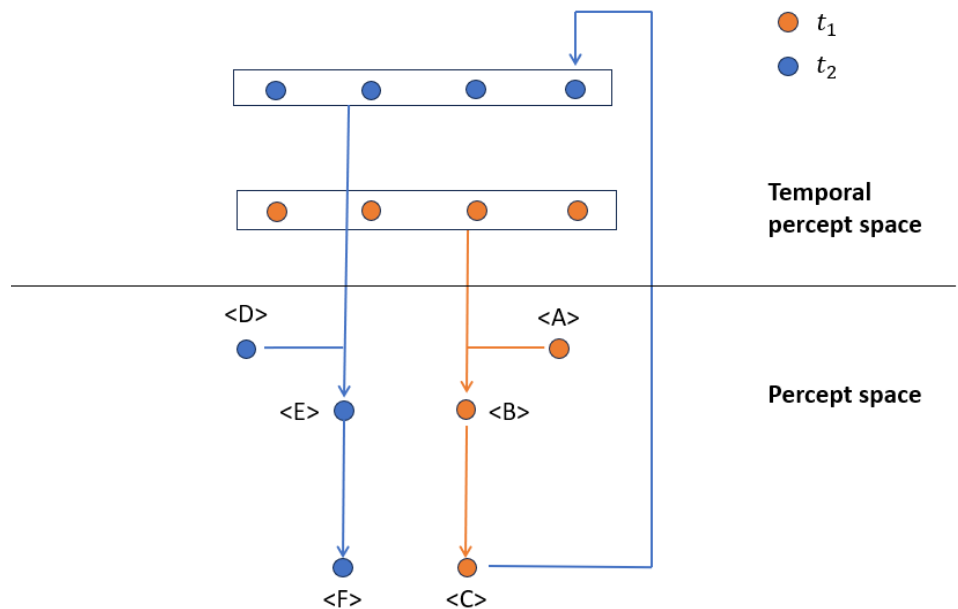

**Figure 4.5:** $S_1$ and $S_2$ are temporal sequences, each with four temporal percepts. The activation of a temporal sequence $S_1$ and <A> at $t_1$, activates the regular percept <B>. Thereafter, <B> activates a closely related percept <C>, kickstarting <C>'s corresponding temporal percepts. After time $t_2 - t_1$ has passed, <C, $t_2 - t_1$> is activated. Coupled with the activation of other temporal percepts within $S_2$, $S_2$ is activated. Thereafter, $S_2$ and <D>, activates <E>, which activates another closely related percept <F> at $t_2$.

The back-and-forth interplay of relational propagation between the percept space and the temporal percept space gives rise to generalizable, complex, and dynamic relational propagation. The following section will include additional real-world examples to showcase the ability of relational propagation to solve problems across different disciplines of knowledge.

## 4.1.4 Problem solving

Problem-solving, at its core, involves producing an appropriate response given a context. In language, problem-solving may refer to generating the most appropriate word given a sequence of words, as in a decoder-only LLM. For mathematics, problem solving may refer to generating an appropriate sequence of steps for derivation. In day-to-day life, problem-solving may involve navigating an optimal path given a location and a destination.

Through repeated exposure to these disciplines, rules may be extracted and consolidated as temporal percepts and regular percepts. Some rules can be "slow", such as the ideas embedded across sentences, the steps within a mathematical formula, or the location within an environment, while others can be "fast", such as the rapidly shifting lexical choices and grammatical adjustments, the numerical changes within each calculation, and limb states and positions. Slow and fast rules, control, and shape the evolution of the response over time, constraining the scope of possible predictions for the next percept of the response. Therefore, effective problem-solving requires learning and consolidating as many rules as possible.

## 4.1.5 Linguistic problem solving

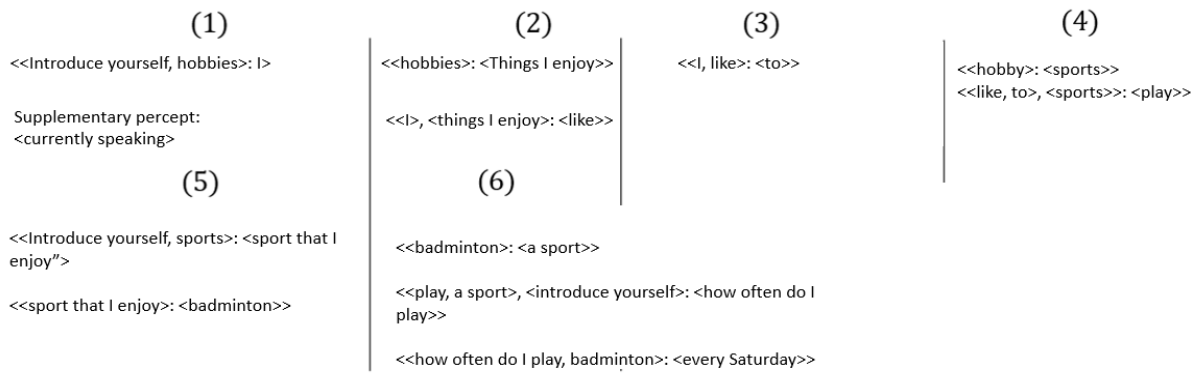

**Figure 4.6:** Temporal percepts in this example are supported by sequence cells. Consider a situation where an individual is asked to introduce themselves, where the goal is to derive the sentence <I like to play badminton every Saturday> with supplementary percepts of <introduction>, <hobby>. **(1)** The percepts of <introduce yourself> and <hobbies> activates <I>, <I> is a reasonable word to start a self-introduction. This may create the supplementary percept of <currently speaking>, to indicate the execution and progression of speech **(2)** <hobby> activates <things I enjoy>, as both are related by definition. Similar to before, activation of related percepts assists in activating relational propagation targets. For example, <hobby> may activate <interest>, which is closely related to <enjoy>, complementing the activation of <enjoy>. Afterward, <I> coupled with <currently speaking>, and <enjoy> activates <like>. **(3)** The temporal percepts of <I, like> activate <to>, possibly consolidated into a temporal sequence due to how often this structure is used. **(4)** A supplementary percept may encode the preconceived decision to discuss sports, <talk about sports in my introduction>, enabling <sports> to be activated. The temporal percept of <like, to> together with the percept of <sports> activates <play>, as <play> is a common verb for sports and physical activities. **(5)** The percepts of <introduce yourself> and <sports>, activates <sports that I enjoy>, which then activates <badminton>, relating to the individual's preferred sport. **(6)** <badminton> activates <a sport>. The temporal percept of <play, a sport> and <introduce yourself> may generate the desire to add an adverbial phrase to complete the sentence. The expectation activates <how often do I play>, which, when combined with <badminton>, activates the final percept <every Saturday>, thereby completing the sentence.

There can be many alternative approaches in addition to the relational propagation steps discussed in **Figure 4.6**. There are a few notable details. First, the percepts of <sports>, <badminton>, and <Saturday> may be conceived prior to the beginning of the sentence, encoded as a complex, summarized percept. It is not strictly necessary for them to be derived and activated only when they are required. Second, many plausible supplementary percepts were omitted for simplicity. Supplementary percepts can provide additional excitation to targets that might not have otherwise been activated. Third, in addition to controlled retrieval, other higher-order functions, such as mental imagery or cognitive control, may play complementary or even foundational roles in problem solving.

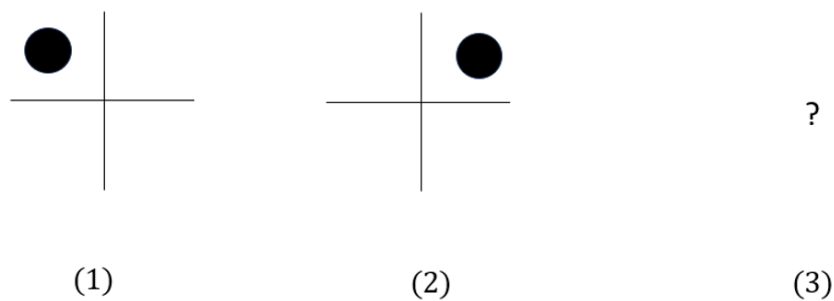

Figure 4.8: A step-by-step demonstration of utilizing a consolidated temporal sequence to infer the possible following item of the pattern question. <<top left>, <top right>: <bottom right>>, is a commonly observed temporal sequence indicative of a clockwise paradigm, likely to be consolidated. <bottom left> is also a valid answer, as the temporal percept sequence of <<top left>, <top right>: <bottom left>>, is also a commonly utilized temporal sequence used during reading or listing.

Alternative approaches for solving the problem in **Figure 4.8** may involve executing a learned visual attention path. For example, the illusion of <square slots> formed by the cross may guide visual attention in a clockwise path. The final point of focus will be the individual's answer.

Additional mechanisms, including executive control, thalamic inhibition, recursive thinking, and decision-making, which will be discussed in subsequent sections, can complement the predictive module. There are questions that cannot be solved with the predictor module alone, examples include mathematical operations ($a < b, a, b \in \mathbb{Z}$) with infinite possible numbers to compare. In the case of infinite question possibilities, relational alone propagation is insufficient, as every number is a distinct percept belonging to a distinct conceptual category. For instance, the question of whether <362> is larger than <363> cannot be answered through relational propagation alone. Other examples include novel or unnatural questions, such as <Can pizza be installed above a car windscreen?>, or abstract questions with analytical depth, such as <Accepting knowledge claims always involves an element of trust?>

These questions heavily rely on executive control to initiate a paraphrase or to stage a counterquestion.

## 4.2 Module 3.2: Prediction module (Episodic sequences)

Proposed anatomical region(s): Hippocampus, distributed cortical regions

The hippocampus is commonly viewed as an indexing system, storing compressed indices that reactivates representations within areas of the neocortex (Teyler & DiScenna, 1986) (Percept space). Partial activation of a subset of neocortical representations corresponding to a memory trace can activate a corresponding population of hippocampal neurons, reinstating the entire memory trace in the neocortex. (Yassa & Reagh, 2013; Teyler & Rudy, 2007) Even with degraded or noisy representations, the whole memory trace can still be recovered; this is a hallmark of pattern completion (Guzowski et al., 2004).

In submodule 4.1, percepts predicted by temporal sequences are decontextualized and general, averaged from many overlapping cortical traces created over learning experiences. Alternatively, submodule 4.2 proposes a model for the recall of contextualized, detailed episodic memories and the generation of dynamic mental imagery.

Episodic memory is canonically viewed as mental time travel (Schacter et al., 2007), in which episodic memories are relived through the sequential replay of events or projected forward to simulate possible futures (episodic future thinking). Nonetheless, others suggest episodic memory should be seen more as an active reconstruction of memory fragments based on cortical schemas (Quiroga, 2020), rather than an accurate replay, citing the presence of illusory details introduced during recall.

We propose that episodic memory may begin as an accurate time-locked sequential replay, where fast percepts quickly degrade due to interference or corruption of episodic memory, while slow percepts are retained for more extended periods of time, at which point reconstruction can compensate for the loss of detail through logical inferencing. Regardless, a replay serves as the blueprint to which reconstruction can be anchored, providing the temporal structure of the episode and the changes in motifs over time. We suggest that the saliency of memories, or the repeated retrieval of memories, can promote long-term storage as an accurate replay in the hippocampus. This view of potentially long-term episodic memory storage is consistent with multiple trace theory (Quiroga, 2020; McKenzie & Eichenbaum, 2011).

This submodule proposes that neurons encoding compressed indices within hippocampal CA3 can be strengthened to each other, and modulated through a precise timing system that controls the temporal transition of indices to reflect coherent replay. This precise timing system that guides the firing of time cells is proposed to be subserved by internal recurrence within the hippocampus (Rolls, 2013) and by entorhinal cortex (EC) ramping inputs (Umbach et al., 2020).

Precise veridical details sustained by replays are likely limited to recent episodic memories, shifting memories towards more schema-driven reconstruction. A few possible reasons for shifting from recall to reconstruction are listed. (1) During retrieval, illusory details from supplementary percepts may lead to the addition of new erroneous information, while failure or deliberate omission can subtract information, generating a new memory trace that competes with the original memory trace. Other factors, including semantic bias, goal-driven reconstruction, or altered experiential context (Nadel & Moscovitch, 1997), can lead to changes in the original memory trace. (2) Recall discriminability degrades over time due to interference from competing traces with overlapping contexts (Yassa & Reagh, 2013). (3) A natural degradation of hippocampal traces through neurogenesis (Kitamura et al., 2009) or synaptic weight decay (Richards & Frankland, 2017). (4) We propose a possible "confirmation bias" effect, a positive feedback loop that biases the retrieval of consolidated schemas. For instance, during the retrieval of an episodic memory, an illusory percept is introduced, creating a new trace. During another controlled retrieval, one uses the illusory detail as a retrieval cue for the memory, under the pretense that the illusory percept is a schema-consistent reminder of the event, and retrieves a new trace consisting of the illusory detail rather than the original trace. This fuels a positive feedback loop and the decontextualization of memories to conform to existing schemas.

## 4.2.1 Nature of episodic sequences

In the view of replay, episodic memory begins as a time-locked sequence of percept contexts, including both temporal percepts and regular percepts. For simplicity, we propose that all attended percepts, whether derived from controlled or automatic retrieval – ATL percepts, sensory percepts, or emotional percepts - contribute to the formation of the episodic memory sequence.

The greater the specificity of the percept, the shorter its temporal persistence and the greater its vulnerability to interference. Raw sensory details degrade first, followed by whole objects, scenes, conceptualizations, and overarching events.

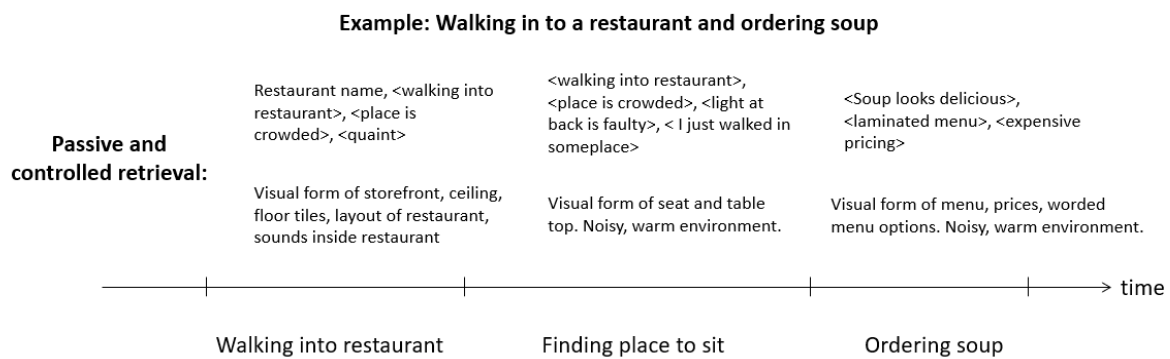

**Figure 4.9:** Percepts that may be captured by the hippocampus through bottom-up perception and top-down programs, during a scenario of walking into a restaurant and ordering soup. This includes conceptual percepts: <place is crowded>, <quaint homely restaurant>, sensory percepts: image of storefront, ceiling, broken lights.

Studies generally indicate that hippocampal time cells provide a general temporal scaffold instead of specific object-time pairings (Ranganath & Hsieh, 2016; Shimbo et al., 2021). Instead, bound co-activation of time cells and perirhinal cortex or CA3 object-encoding cells encodes the conjunction of objects and time, forming the substrate of temporal percepts. A proposal for percepts and temporal percepts is illustrated in **Figure 4.11**, where temporal percepts can point to a regular percept, which is then integrated with the firing of time cells to become a temporal percept.

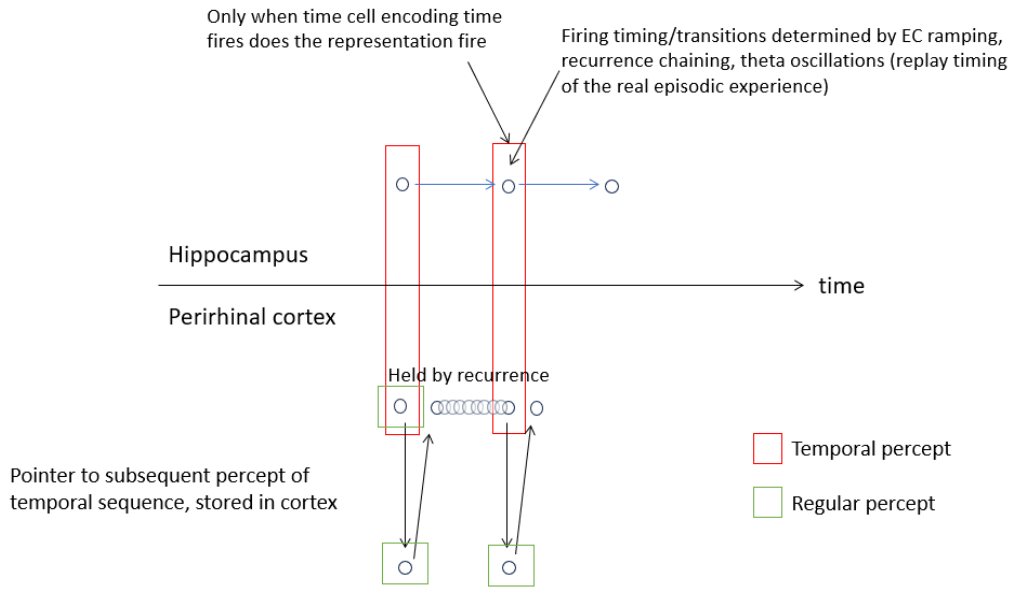

**Figure 4.10:** An illustration of a possible mechanism for the replay of an episodic sequence. Hippocampal time cells provide temporal structure to percepts by ensuring that subsequent regular percepts of the episodic sequence may activate only when paired with particular time cell activations, mimicking the behavior of cortical temporal percepts and temporal sequences. Percepts may be temporarily sustained in perirhinal recurrent networks, awaiting the activation of corresponding time cells.

Temporal percepts fluctuate across percept contexts within episodic sequence as new time cells activate and while prior time cells cease. On the contrary, regular percepts can persist across time, as illustrated in **Figure 4.12**, awaiting their corresponding time-cell activation. We identify two key steps for the retrieval of episodic sequences.

1. The auto-association of any percept context within the episodic sequence, through the matching of percept with percept contexts within the sequence. The provision of more percepts, whether through relational propagation of external perceptions, can increase the accuracy and likelihood of retrieving the desired memory (**Figure 4.12**), analogous to a retrieval system.
2. Upon the activation of any percept context, successive percept contexts activate in sequence, reflecting the temporal structure of the memory. For example, percept context A may activate first, followed by percept context B one second after, and percept context C two seconds after.

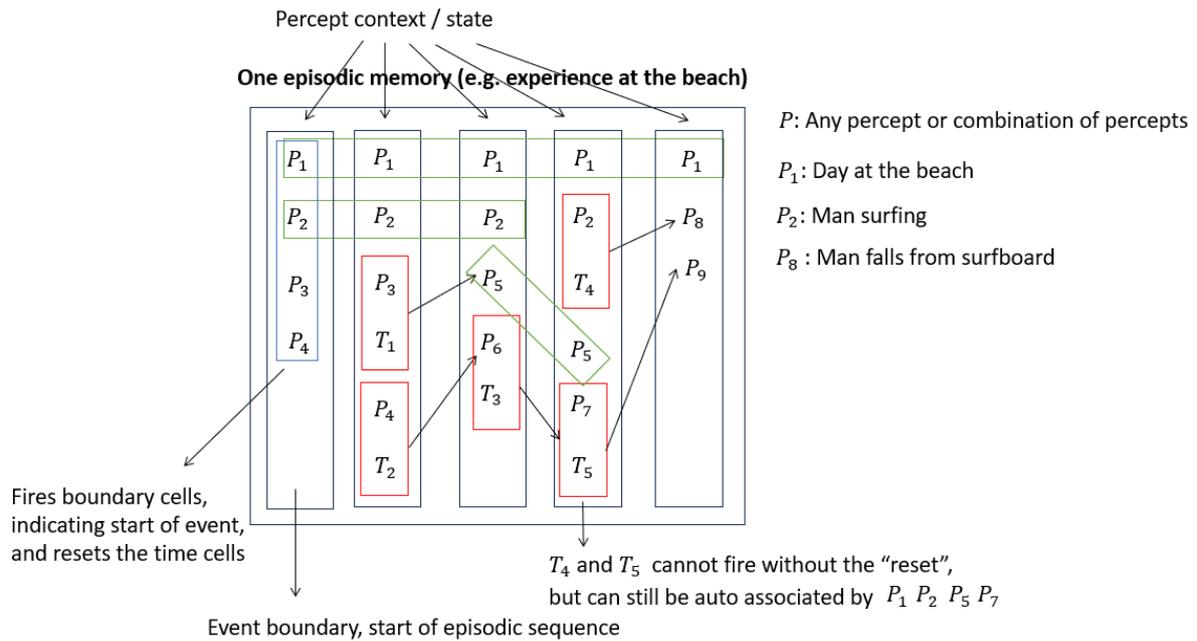

**Figure 4.11:** Diagram depicting an episodic memory of an individual's experience at the beach. Annotations provide suggestions for what the enumerated percepts may represent. Slow percepts are usually the overarching descriptions of the episodic event, while fast percepts are the rapidly changing objects and scenes observed. The transition from $P_2$ to $P_8$ may indicate a change to the scene, perhaps capturing the moment a surfer is thrown off the board.

Suppose an individual is trying to recall the experience of listening to a song while jogging on a particular day. He may retrieve the transmodal percept of <jogging>, the title of the song he was listening to, the atmosphere, the weather while jogging, and the emotion he feels upon hearing the song. With sufficient percepts, the percept context may be auto-associated, supplementing the remaining percept details.

Likewise, consider an individual trying to imagine what he had for breakfast two days ago, he would retrieve transmodal percepts of <breakfast>, <my dining table (vis)>, or the relevant experiences two days ago, such as <places that I were>, <sights seen that day>, which may activate more relevant percepts through relational propagation.

Temporal percepts can also be activated through controlled mental imagery. We propose that the initiation of controlled imagery may constitute a salient event boundary that resets hippocampal time cells, enabling the tracking of the new temporal sequence. For instance, to recall the lyrics of a song, an individual may hum or imagine the melody, matching the actual episodic sequence over time.

Although any percept context can be auto-associated through sufficient supplementary percepts, unexpected or novel experiences may contain infrequently observed percepts that are less likely to be inferred through relational propagation. These experiences tend to be orthogonalized from other episodic memories, reducing their susceptibility to interference and increasing the likelihood of accurate retrieval upon the activation of those infrequent percepts.

# 5. Module 4: Executive control module

Corresponding anatomical region(s): prefrontal cortex (PFC)

The prefrontal cortex (PFC) is the anterior section of the frontal lobe responsible for higher-order cognitive functions. The PFC exhibits dense recurrent connectivity, capable of maintaining persistent activity to support working memory (Curtis & Sprague, 2021).

Module 4 is organized into six submodules: rostral-caudal feedforward network, action selection module, fast buffer, relation store, reward network, and meta controller, each corresponding to a distinct anatomical region. The subsequent sections will examine each of these submodules in the order presented.

It should be noted that each submodule is intended not as a precise biological description of its corresponding anatomical region, but rather as a simplified modular proposal of its putative function. In some cases, this abstraction may be biologically inaccurate and overly reductive.

## 5.1 Module 4.1: Rostral-caudal feedforward
Corresponding anatomical region(s): Rostral PFC, Dorsal PFC, SMA, M1

We describe the prefrontal rostral-caudal hierarchy (Rostral PFC → Dorsal PFC → Premotor/supplementary motor area (SMA) → Motor Cortex (M1)) as a feedforward network, receiving percept contexts sustained in working memory as inputs, and generating internal and external actions as outputs. Internal actions encompass top-down biases, controlled retrieval, working memory control, etc., while external actions are motor behaviours and movements. The feedforward network will be termed the feedforward action network (FFA) and is conceptually similar to a reinforcement learning (RL) policy network with memory, implemented with recurrent networks.

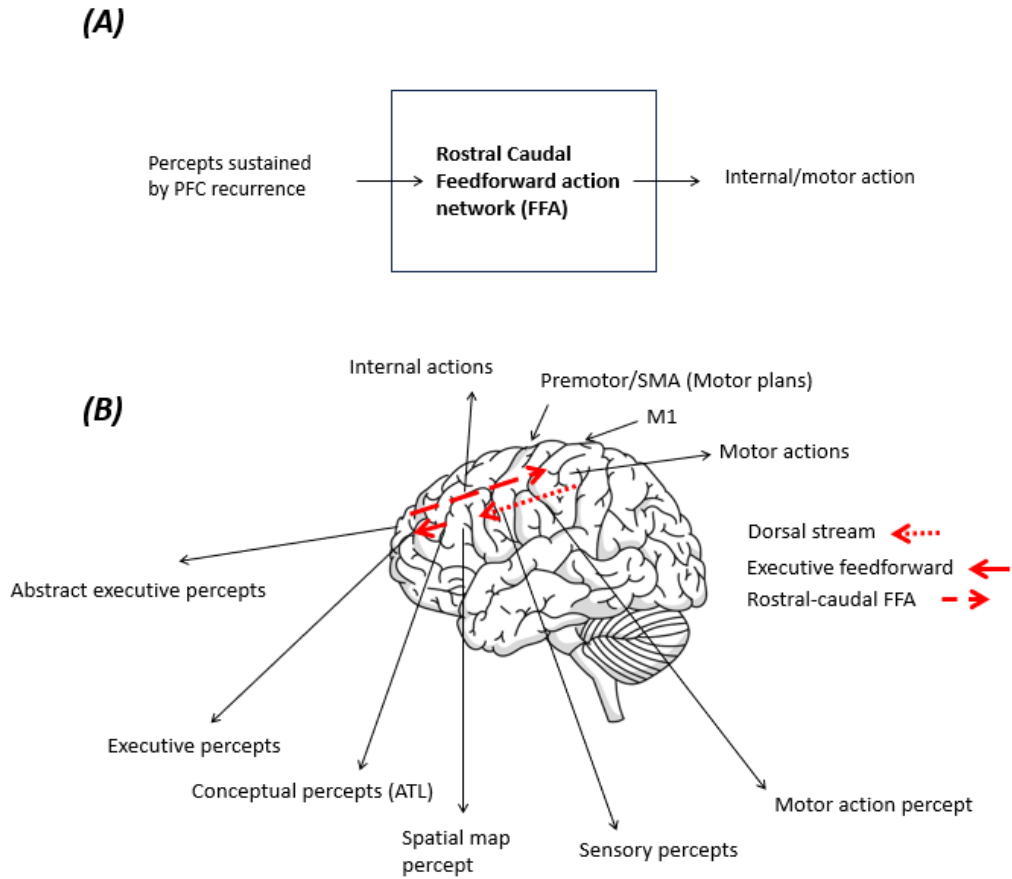

**Figure 5.1: (A)** Schematic of the FFA, where inputs consist of percepts maintained by recurrent prefrontal networks, and outputs correspond to both internal and external motor actions. **(B)** Diagram of the brain indicating general locations of the FFA, dorsal stream, and executive feedforward network, along the six hierarchical levels of percepts, with percepts of lower abstraction levels input to increasingly posterior areas of the FFA.

FFA is modulated by dopamine, which acts as a teaching signal that encodes reward prediction errors (RPEs) (Schultz, 2016; Glimcher, 2011). A higher-than-expected reward causes a spike in dopamine, while a lower-than-expected reward causes a drop in dopamine from the tonic baseline. Dopaminergic (DA) neurons in the ventral tegmental area (VTA) and substantia nigra pars compacta (SnC) project to the striatum, prefrontal cortex, and hippocampus through DA pathways. Following RPE signalling, dopamine diffuses to the active synapse via volume transmission, inducing LTP of previously activated synapses marked by short-term synaptic traces (Shindou et al., 2019). Conversely, during a drop in baseline dopamine, the neuron is biased towards LTD.

The FFA maps percept contexts to actions. Context-action associations that evoke dopaminergic responses – signalling a higher-than-expected reward – are preferentially strengthened based on the assumption that the executed action was more favourable than predicted. The strengthening of context-action associations during dopamine underlies incentive salience, as reflected in an increased tendency to repeat the same action in similar future contexts to attain higher rewards. Additional details on reward mechanisms are discussed in Module 4.5.

We propose a hierarchical structure within the FFA comprising six levels of perceptual abstraction, each feeding into different levels of the feedforward network, topologically corresponding to increasingly posterior regions of the PFC. Note that percepts can include both regular and temporal percepts.

First, abstract executive percepts reside at the highest end of the hierarchy, situated in anterior regions of the PFC. Abstract executive percepts are metacognitive – prospective memory, multi-tasking, confidence, familiarity, "Aha moments", retrieval failure and forgetting, and time estimation – are processes that require neural oversight and are integrated over time. Instead of relying on information encoded by neurons, oversight arises from the integration of signal-flow directions, regional activity levels, and other neural indicators. For example, one may be aware of their lack of knowledge about a question if controlled retrieval precedes the absence of a response or an increase in activity in the anterior cingulate cortex (ACC), indicating a mismatch or failure (Braver et al., 2001). Time estimation, for example, could be oversight of internal oscillations, such as striatal neurons firing in rhythmic patterns (Jin et al., 2009). We propose that abstract executive percepts are outputted from the executive feedforward network (EFF), where the inputs to the EFF are executive percepts.

The second level consists of executive percepts. Similar to abstract executive percepts, executive percepts require neural oversight, involving the integration of endogenous neural signals. In contrast to abstract executive percepts, we propose that executive percepts integrate information over shorter time intervals and with reduced complexity. Executive percepts may include immediate indicators representing the updating or protection of working memory, direction of flow of endogenous signals, retrieval indicators, mismatch failures, gating, and ungating signals. This is contrasted with abstract executive percepts, which may encode combinations of executive percepts over time to capture more complex information about the execution status of multiple tasks, complex temporal estimations, or prospective memory. For example, determining the status of controlled retrieval necessitates the sequential monitoring of executive percepts relating to the transfer, retrieval, region activity, and error signals over time, while multitasking requires the concurrent tracking of multiple action sequences and close monitoring of stages of ongoing actions.

The third level consists of conceptual percepts originating from the ATL. The fourth level consists of cognitive maps derived from the parietal cortex, hippocampus, and higher sensory areas. The fifth level includes percepts from mid- to low-level sensory areas, including auditory, visual, or proprioceptive information. Lastly, the sixth level consists of motor percepts encoding motor firing patterns in the motor cortex.

The hierarchical level of percepts within the network correlates with the slowness of percepts, with higher hierarchical percepts usually being slower percepts. This ranges from fast-changing proprioceptive information about limb positions to slower-changing spatial environments, to conceptual information, to executive percepts encoding current task sequences, to abstract executive percepts encoding multiple task sequences.

The FFA differs from hierarchical reinforcement learning (HRL). HRL architectures describe a hierarchical cascade of control, where high-level controllers provide goals that direct low-level policies (Nachum et al., 2018). The FFA proposed is a monolithic policy network, where all percepts, regardless of abstraction level, are conjunctively used to generate the action. The derivation of behavioural options or subgoals in a specific situation is not facilitated by interactions between goal controllers, but rather by relational propagation within the temporal and percept spaces. For instance, the goal of <cooking noodles> may associate to the subgoals of <boiling water> or <pouring seasoning>, depending on the results of relational propagation when provided with context-relevant supplementary percepts. The subgoal of <pouring seasoning> may be activated after <water has boiled>, while <boiling water> may be activated after the thought of <cooking instant noodles> comes to mind.

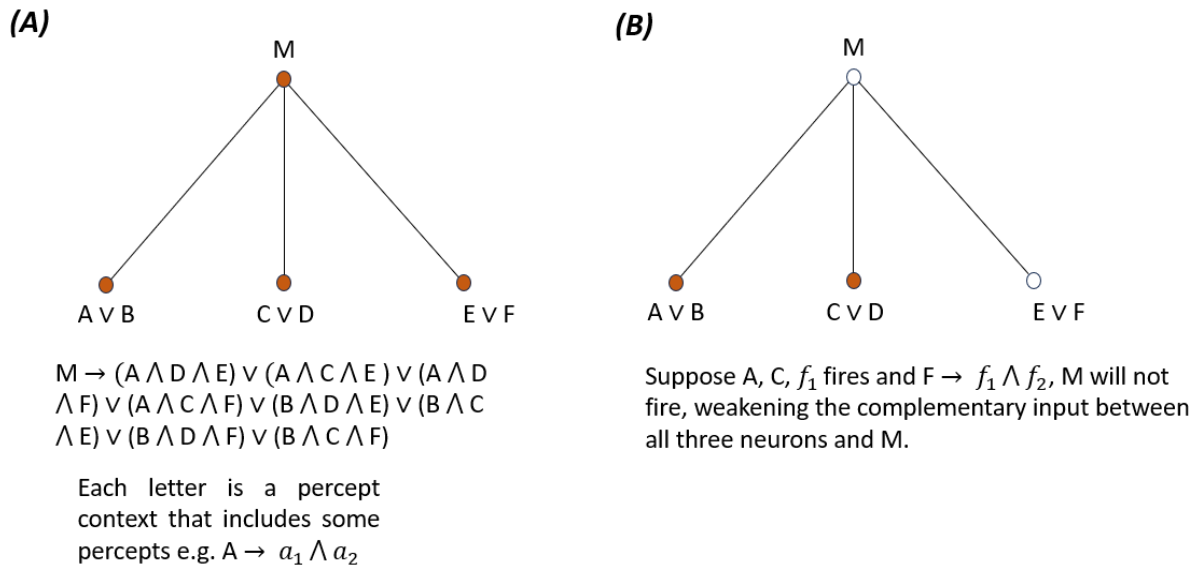

**(A)**

M → (A ∧ D ∧ E) ∨ (A ∧ C ∧ E) ∨ (A ∧ D ∧ F) ∨ (A ∧ C ∧ F) ∨ (B ∧ D ∧ E) ∨ (B ∧ C ∧ E) ∨ (B ∧ D ∧ F) ∨ (B ∧ C ∧ F)

Each letter is a percept context that includes some percepts e.g. A → $a_1 ∧ a_2$

**(B)**

Suppose A, C, $f_1$ fires and F → $f_1 ∧ f_2$, M will not fire, weakening the complementary input between all three neurons and M.

**Figure 5.2:** Each letter corresponds to a percept context consisting of multiple percepts, for instance, the percept context $A$ encompasses the percepts of $a_1$ and $a_2$, where $A$ will be activated only if both $a_1$ and $a_2$ are activated. The target neuron M activates only when all three input neurons are simultaneously active, where the input neuron may activate when one of the percept contexts is activated. **(A)** While multiple combinations of percept contexts can activate M, each combination requires at least three percept contexts to be activated. The combinations required for M's activation are more specific than the input neurons' combinations, which only require two percept contexts to be activated. This demonstrates a key property of mixed selective neurons. Mixed selective neurons activate for a larger number of percept combinations, but require more percepts to activate for each combination, as compared to their mixed selective inputs. **(B)** The absence of any percept required for M to activate can lead to weakening of the complementary input between the three input neurons and M. Complementary input weights in mixed selective neurons are less stable than those in percept selective neurons, which can converge on a stable prototype. Feature-selective neurons in the VPS respond to individual objects or single features. When a complementary input activates but the target neuron does not, the complementary input becomes weakened, suggesting that the feature it encodes is likely absent within the feature encoded by the target neuron. As each complementary input only represents one feature, the margin of error in this inference is minimal. Conversely, mixed selective neurons respond to multiple combinations of percepts. When a complementary input activates but the target neuron does not, the complementary input is weakened. Nonetheless, the precise reason is uncertain, as it could indicate the absence of $f_1$ or the absence of $f_2$ or a combination thereof. This results in a broader margin of error during learning, as multiple percept combinations may be implicated.

Neurons in the PFC are known for being mixed selective, responding to multiple combinations of task variables (Ramirez-Cardenas & Viswanathan, 2016), where distributed populations are capable of encoding all possible task information (Rigotti et al., 2013).

Demonstrated in **Figure 5.2A,** complementary inputs created by mixed selective neurons are more unstable as compared to complementary inputs in the VPS. Mixed selectivity neurons offer computational advantages, namely flexibility and extensive generalization capabilities curated through many learning experiences. The protracted development timeline of the PFC, in contrast to the fast maturation of visual cortex neurons (Oztop et al., 2013), may be a direct consequence of this difference.

Across feedforward layers of the FFA, mixed selective neurons encode a growing number of increasingly specific percept contexts. While individual neurons may respond to multiple combinations of percepts, a specific combination of percepts can be decoded through populations of neurons. This enables the FFA to represent every possible combination of percepts — goals, motor plans, rules, and environments — and to perform a total mapping onto corresponding actions.

## 5.1.1 Motor plans and actions

Motor actions percepts in the motor cortex constitute the final stage of the FFA network. At this stage, mixed selective neurons encode highly specific combinations of percepts. These percept combinations represent goals, motor plans, rules, and environments, which are integrated with previous motor states, to guide subsequent actions. The process is illustrated in **Figure 5.3**.

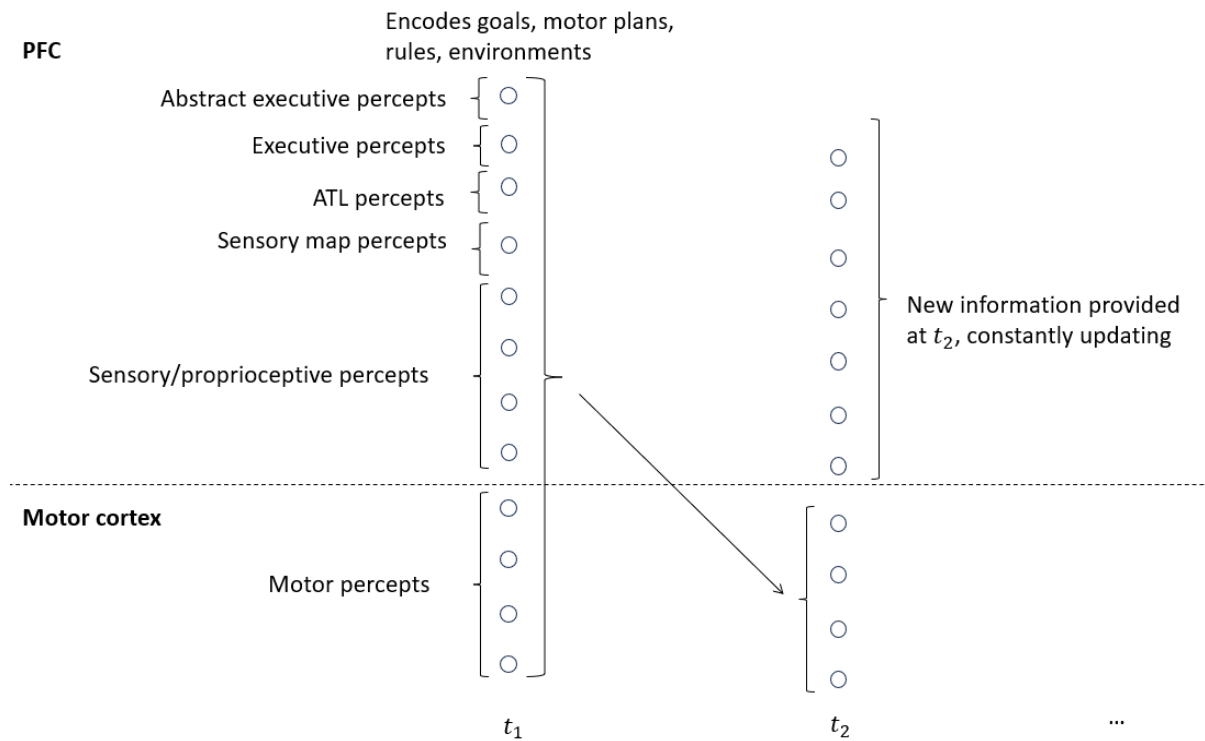

**Figure 5.3:** An illustration of the final stage of FFA in the motor cortex. Motor neurons are linked to specific actions. Percepts from earlier layers of the FFA have been integrated such that neuron population activity decodes specific motor plans and goals.

## 5.2 Module 4.2: Action selection module
Corresponding anatomical region(s): cortical-basal-ganglia-thalamus loop (CBGT)

The basal ganglia exert tonic inhibition over thalamic targets. Through the direct pathway, disinhibition of the thalamus permits excitatory thalamocortical drive, enabling the execution of voluntary motor commands (Rocha et al., 2023). Strong lateral inhibition within the striatum has been proposed to give rise to WTA mechanisms (Kropotov & Etlinger, 1999), through decorrelation of neural activity. The indirect and direct pathways of the basal ganglia play complementary roles. The direct pathway selects the desired movement, while the indirect pathway suppresses competing movements (Tecuapetla et al., 2014).

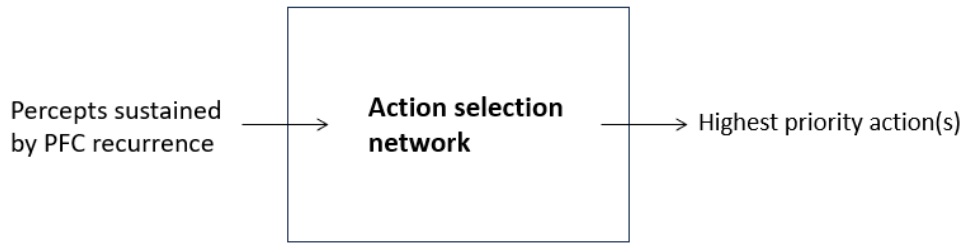

**Figure 5.4:** Schematic of the action selection network, with inputs being percepts, and output being the highest priority action(s).

We propose that the CBGT may be modeled as an action selection network, where inputs are percept contexts and the output is the action(s) of the highest priority. The action selection network complements the FFA by facilitating direct suppression of low-priority actions, amplification and stabilization of higher-priority actions, and the execution of corresponding actions. The action selection network plays a crucial role in supporting the FFA.

Firstly, the FFA operates based on a "matching principle", whereby the probability of an action's execution increases with the number of percepts aligning with the action's preferred percept context. Consequently, weakening of different context-action pairs requires multiple trial-and-error learning experiences. For example, the percept context encompassing the <do not move!> command should suppress all action-related percepts, regardless of whether they have previously co-occurred with the command. However, it is impractical for every possible action to be jointly experienced with the command. The action selection network compensates for this limitation by providing a mechanism for generalized inhibition across actions, functioning as a binary control signal. The action selection network enables the <do not move!> command to induce generalized suppression of movement when activated, independent of the specific context or supplementary percepts. Secondly, the FFA does not facilitate action selection. When multiple actions are referenced through percept contexts, they cannot be executed simultaneously, necessitating the prioritization of desired actions and suppression of less relevant ones. Third, the action selection network is critical for habit suppression. Repeated associations between a percept context and a specific action give rise to a habit, which is difficult to override without control mechanisms provided by the action selection network.

We propose that the action selection network may be conceptualized as a fast, versatile system that can rapidly adjust relational weights between contexts and actions, while the FFA is a slow, gradual system that encodes generalized relationships between contexts and actions.

## 5.2.1 Development of FFA and PFC control

Motor babbling, reflexes, and mimicry may serve as the primary means by which an infant may spontaneously perform actions without prior experience of explicit motivations. Mimicry can occur through the imitation of another person's actions or the motion of external objects, facilitated by the activity of mirror cells (Oztop et al., 2013)

During motor babble, the execution of actions represented by motor percept sequences is accompanied by concurrent PFC percept representations. The co-occurrence of PFC percepts and motor percepts leads to the strengthening of their connections, resulting in the creation of a relational prototype between PFC percepts and motor percepts, and the possibility for future occurrences of those PFC percepts to activate those motor percepts. This mechanism allows PFC percepts to develop relations to motor percepts in the motor cortex, enabling the execution of desired actions during their reactivation. Thereafter, repeated activation of motor percepts by PFC percepts further strengthens connections between them.

According to this hypothesis, the PFC encodes the label of the visual motion percepts observed during motor babble — what the movement looked like — which becomes the trigger for involuntary actions. Over time, the context-action relations are refined by rewards, transforming involuntary actions into voluntary ones. Mimicking adults performing positive behaviors can accelerate this transition, as these learned actions are likely already examples of reward-seeking behaviors.

The CBGT network operates in consonance with the FFA to facilitate percept-to-action strengthening through repetition and reward. Repetition and reward contribute both to the reinforcement of connections within the FFA and the disinhibition of thalamic targets for the corresponding percepts in the CBGT. While their exact roles may differ, both lead to an increased probability of action execution during strengthening and a diminished probability of action execution during weakening.

## 5.3 Module 4.3: Relation store and fast buffer

Corresponding anatomical region(s): DLPFC (Relation store), VLPFC (Fast Buffer)

Module 4.3 postulates the existence of a relation store that temporarily maintains associations between two or more percept contexts. Linking two percept contexts replicates the general cue-command form of rules. We propose that storage within the relation store requires precise apprehension of the rule in its exact cue-command form. For example, if a reward occurs after pressing a button in response to a light flicker, the explicit apprehension of <light flicker> percept followed by <press the button> percept is necessary for the storage of the rule into the relation store.

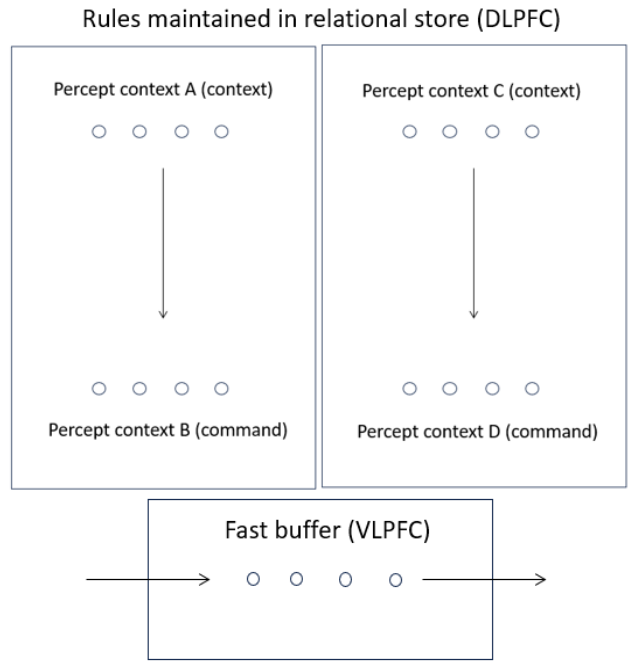

**Figure 5.5:** A simplified diagram showcasing the storage of rules – represented as an association between cue percept and command percept, in the relation store (DLPFC). The fast buffer's (VLPFC) role complements the relational store by controlling the transition between percept contexts.

We propose that storage of relations within the relational store involves sequential ungating of the relational store to represent the cue percept, followed by the representation of the command percept after some delay. Sequential activation of percepts to quickly and temporarily relate two percepts together may be facilitated by short-term plasticity (STP) (Fujisawa et al., 2008).

The precise switching of percepts entering the relation store is mediated by an additional subsystem, the fast buffer (VLPFC), which dynamically transitions between context percept and command percept by suppressing the cue percept and exciting the command percept at appropriate moments, complementing the relational store's function. Both the fast buffer and the relational store are subject to local inhibition and modulation by the action selection network.

When perceived percepts match the cue percept stored in the relational store, the corresponding command percept is activated and serves as an input to the FFA. Thereafter, the FFA associates the command with its corresponding action, thereby executing the command.

## 5.4 Module 4.4: Reward prediction network

Corresponding anatomical region(s): VMPFC, OFC

The VMPFC and OFC are central components in the brain's reward network (Ullsperger et al., 2014; Howard et al., 2015). The VMPFC and dorsal OFC receive inputs from regions such as the hippocampus and amygdala (Haber, 2011), enabling them to generate signals for predicted rewards. The ventral OFC receives predominant inputs from sensory areas

(Rolls, 2004) to represent received reward signals. In this module, we model the VMPFC as a reward-prediction network, while the ventral OFC as a received-reward network.

Inputs to the reward prediction network include the current and predicted percept contexts, while inputs to the received reward network include only the current percept context. In addition to percepts, inputs to both networks can include internal bodily states such as hunger, stress, or exhaustion. Neurons in both networks possess mixed selective properties, enabling them to encode numerous specific combinations of context reward pairs. Likewise, population activity encodes the predicted and received reward values.

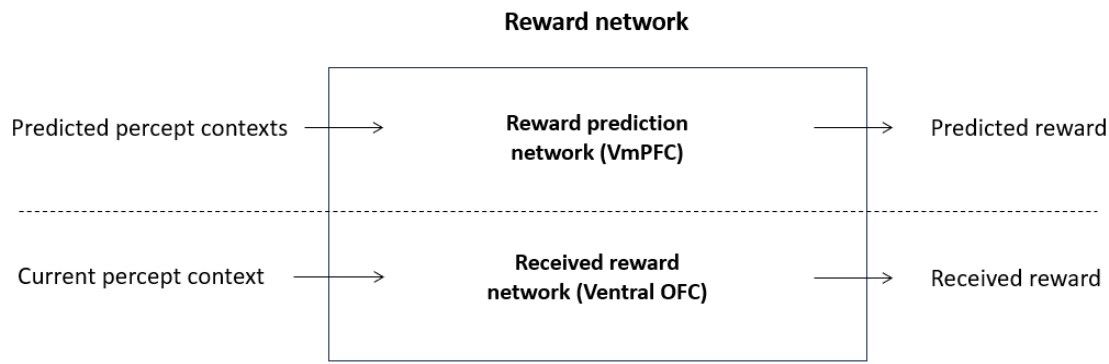

**Figure 5.6:** Schematic of the reward prediction network and the received reward network. The reward prediction network uses predicted percepts from the prediction module to generate reward predictions. In contrast, the received-reward network does not receive predicted percepts and instead generates sensory-driven, immediate rewards.

A network where the input is the current state and the output is reward is similar to an reinforcement learning value function. Indeed, the brain reward system is commonly paralleled to a temporal difference (TD) reinforcement learning model (Schultz et al., 1997). In TD models, the network learns and updates its reward predictions based on temporal difference errors.

$$\text{TD error} = R_{t+1} + \gamma V(S_{t+1}) - V(S_t) \text{ (Montague et al., 1996)}$$

Where, $S_t$ and $S_{t+1}$ are the current and subsequent states, respectively, $V(S)$ is the predicted reward, and $R$ is the received reward.

Unlike canonical model-based TD models that bootstrap value estimates to successive states, the brain can generate predictive sequences that simulate rollouts through the prediction module. Weights between percepts reflect state transition probabilities. For instance, percepts with strong asymmetrical weights are likely to occur that order, while percepts with weak asymmetrical weights to the regular percept is less likely to occur in that order. Well-conditioned cues induce higher weights between percepts, while one-time events or remotely learned associations induce lower weights.

In the semantic module, the ATL retrieves percepts based on general decontextualized relations. In the generalized prediction module, the temporal percept space encodes general decontextualized sequences. In the episodic prediction module, the hippocampus stores contextualized time-locked sequences of percept contexts. Combining three modules can generalize across many novel states while providing diverse predictions of future percept contexts, effectively simulating a rollout over all probable future percept contexts, which are then input to the reward prediction network.

This hypothesis is aligned with many real-world observations of dopamine (encoding RPE), such as phasic dopamine scaling with expected value, magnitude of reward and probability of reward (Schultz, 2010), dopamine shifting its firing to the cue after cue conditioning (Starkweather et al., 2017), decline from baseline dopamine when the reward is not delivered at expected time (Niv et al., 2005), rapid dopamine transients and fluctuations that reflect highly dynamic RPE adjustments.

Suppose the <bell ringing (aud)> is conditioned to <sugar (rwd)> reward delivered five seconds later, implying a high weight between <bell ringing (aud)> and <sugar (rwd)>. When <bell ringing (aud)> activates unexpectedly, <sugar (rwd)> is activated, inputted into the reward prediction network, and mapped to its corresponding reward value. The change in predicted reward triggers a dopamine transient, subsequently returning to baseline without further change in predicted reward. Dopamine may spike again afterward, due to the supplementary percepts observed after <bell ringing>, which may change the probability or magnitude of predicted reward.

If the reward exceeds the expected duration of reward or surpasses the expected reward magnitude, another dopamine spike will be generated (Deng et al., 2023). Conversely, if a reward fails to arrive at the expected time, dopamine levels drop below baseline, as the predicted reward decreases. For extended or delayed rewards, it is helpful to consider multiple "timestamps" representing expected reward moments (**Figure 5.7**).

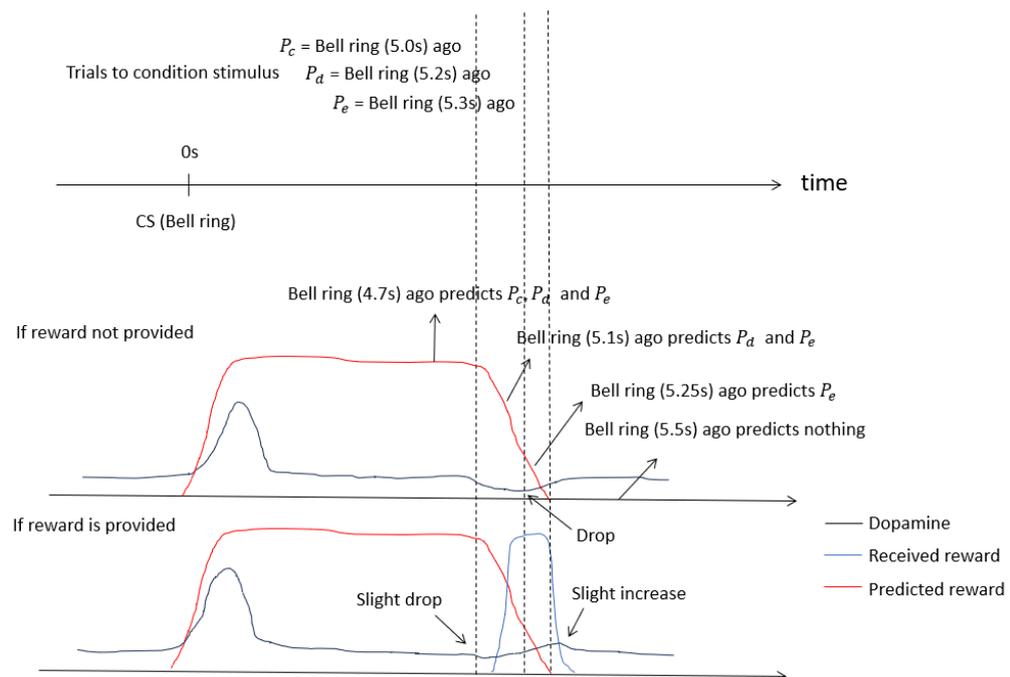

**Figure 5.7:** An illustration of the dynamics of dopamine, received reward, and predicted reward. Consider that the conditioning of <bell ringing (aud)> to <sugar (rwd)> is developed through three separate trials, with slight variability of timing of <sugar (rwd)> delivery during each trial. The graphs below illustrate a theoretical progression of reward signals in a scenario where reward is delivered but delayed, and in one where it is not. In both cases, there is a drop

in predicted reward at the exact timings corresponding to reward delivery during conditioning trials. When the reward is omitted, dopamine levels drop below baseline, producing a negative RPE. Conversely, when the reward is provided but delayed, a slight spike in dopamine is observed as the actual reward exceeds the expected duration of the reward.

RPE arises when the reward prediction network inaccurately estimates the reward magnitude for a given percept context. Dopamine is then released to adjust future reward predictions for erroneous percept contexts, through mechanisms described below.

(1) Synaptic tagging has been proposed to be the substrate for solving the distal reward problem (Päpper et al., 2011; Izhikevich, 2007), where the firing patterns of neurons for prior states are absent during the reward arrival. Through synaptic tagging, erroneous percept contexts encountered before rewards can be targeted by dopamine for strengthening to higher rewards, such that future encounters with the erroneous percept contexts will predict a higher reward than before.

(2) In the hippocampus, dopamine promotes synaptic plasticity, supporting the encoding of episodic memory during salient reward situations (Du et al., 2016). where the erroneous percept context occurring shortly before the RPE is strengthened to the percept contexts after the RPE, future encounters with erroneous percept contexts will generate more accurate percept context predictions and consequently a more accurate reward prediction. For remote or one-time events that are not reinforced, the newly formed relations of the erroneous percept context will likely decay due to interference.

(3) Tagged synapses in the FFA representing actions performed during erroneous percept contexts are strengthened during elevated dopamine. The brain assumes that actions performed within erroneous percept contexts may have led to the generation of the unexpected reward. The connections between erroneous percept contexts and corresponding actions that were performed are strengthened, such that future encounters with the erroneous percept context or similar percept contexts are more likely to execute their corresponding actions. Over time, the weight between a particular percept context and its corresponding actions should converge on the actual reward values received when performing the action within that context.

(4) Dopamine strengthens striatal pathway connections (Calabresi et al., 2007) between erroneous percept contexts and their corresponding actions, increasing their priority for execution. This means that future encounters with erroneous percept contexts are more likely to execute their corresponding actions.

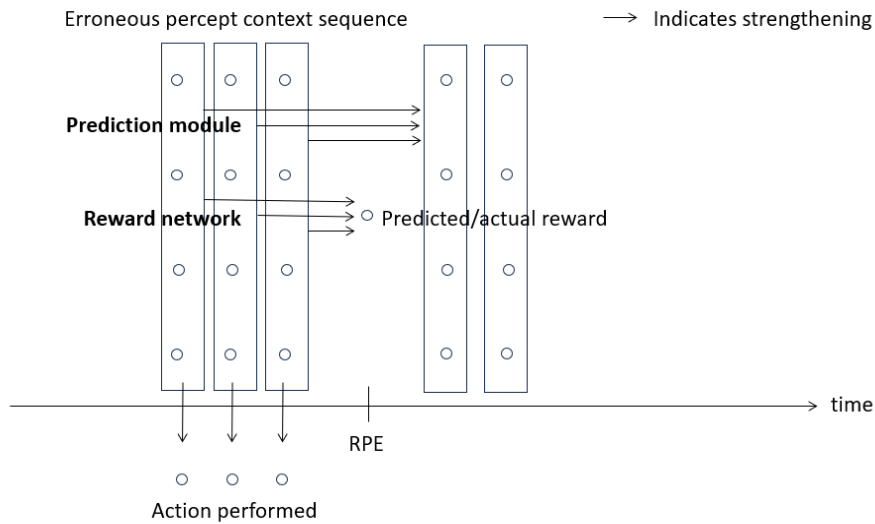

**Figure 5.8:** An illustration of the first three mechanisms of dopamine discussed in the main text, including (1) strengthening between erroneous percept contexts and the predicted reward, (2) strengthening between erroneous percept contexts to percept contexts observed after RPE, and (3) strengthening between erroneous percept contexts and actions that were performed during those contexts.

## 5.5 Module 4.5: Abstract executive store, EFF

Corresponding anatomical region(s): Rostral PFC

The rostral PFC is the most anterior part of the PFC, often associated with functions such as prospective memory, multitasking (Volle et al., 2011), metacognition (Baird et al., 2013), and reality monitoring (differentiating real from imagined stimuli) (Turner et al., 2008). The rostral PFC is located at the front of the FFA. We propose that the rostral PFC encompasses the EFF while simultaneously serving as a store for the maintenance and derivation of abstract executive percepts, thereby complementing the EFF.

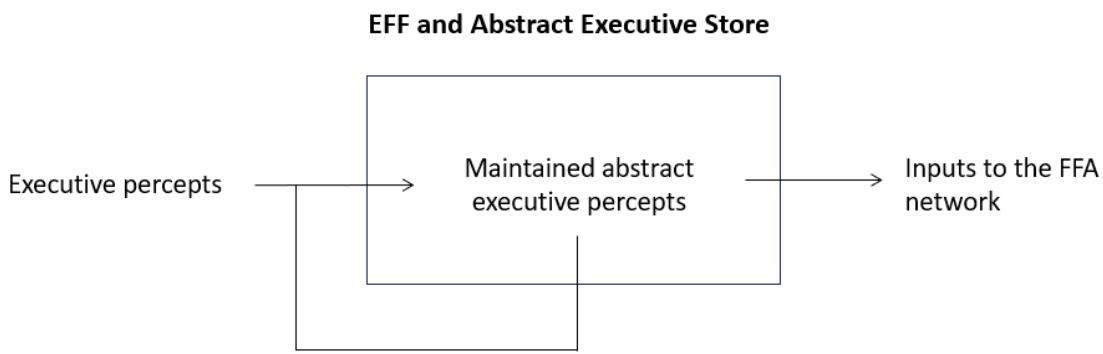

**Figure 5.9:** Schematic of the abstract executive store and EFF. Inputs to the EFF are executive percepts, and outputs are abstract executive percepts, which are maintained in the abstract executive store.

This is the final module, completing the model of human cognition. The following sections will discuss real-world examples and examine how these modules can interact to solve a wide range of tasks.

# 6. Real-world examples

This section aims to demonstrate how the model of human cognition may be implemented, extended, and utilized to solve a wide range of tasks. This section is divided into four subsections, including recursive thinking, decision-making, meta-learning, and the formalization of neural programs.

## 6.1 Recursive thinking

Recursive thinking (RT) is a process that involves the re-representation of a stimulus, granting problem-solving capabilities that would otherwise have been impossible. RT can either be external, through self-initiated motor actions, or internal, through mental imagery or controlled representations. Conceptually, recursive thinking is similar to a chain-of-thought, where articulated steps are perceived and used for further articulation, until the answer is attained.

Recall that the relational propagation may be used for generalization across novel experiences. However, in disciplines such as mathematics, where accuracy is essential, number percepts are discrete objects that do not generalize. Consequently, generalization through relational propagation is insufficient: the consolidated temporal sequence of <<1, +, 1, =>: <2>> cannot be generalized to evaluate <2, +, 2, =>. Instead, another temporal sequence, <<2, +, 2, =>: <4>> needs to be learnt independently to derive the answer. Similarly, comparing two arbitrary numbers <a > b> would require infinitely many temporal sequences. This presents a fundamental problem for knowledge disciplines with infinite variations in problems and high-accuracy requirements, where RT is mandatory.

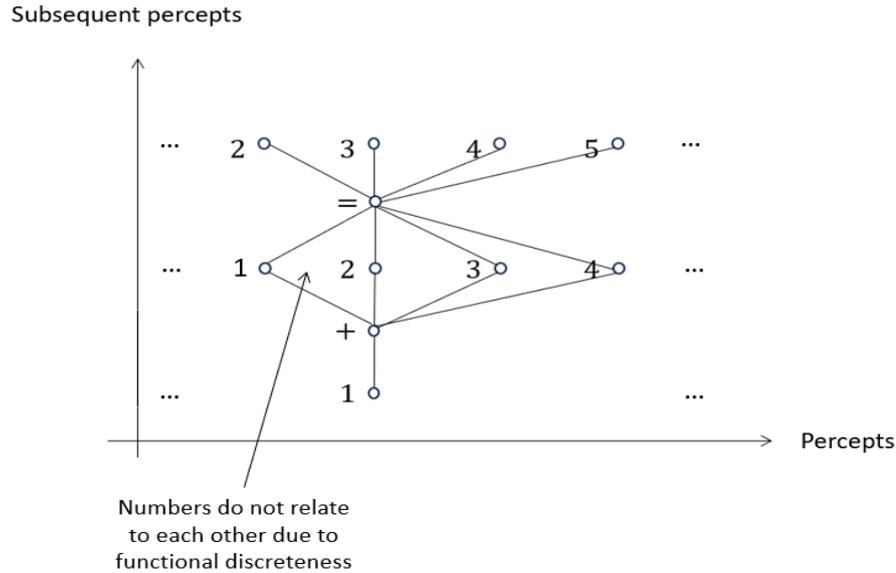

**Figure 6.1:** An illustration demonstrating temporal sequences of addition expressions. The lack of relational meaning between numbers suggests that every "addition" temporal sequence needs to be learned separately.

To clarify the role of RT, several prominent mechanisms along with their relevant sensory modalities have been identified. While some functions are specific to a particular sensory modality, others may be transmodal. Each mechanism will be substantiated with examples.

## Internal (imagery/attention)

1. Restructuring or changing the temporal order of stimuli perceived through attending to particular regions of stimulus in a sequential order. (Visual/auditory).
2. Reformulating or paraphrasing the main problem into sub-problems, such that the solutions to these sub-problems collectively address the original problem. (Auditory).
3. Problem solving through general recursive perception (All sensory channels)

## External (motor action)

4. Dynamic patterns of movement created by actions may elucidate conceptual ideas when observed from third-person perspectives, such as in the game of *charades.*
5. Drawing, writing, or labelling may enhance visualization of problem premises, enabling the derivation of concepts, methodology, or solutions through observation.

## Examples for (1)

The first mechanism involves the use of imagery for RT. When we perform mental arithmetic, such as multiplying or adding two numbers together, we may utilize the addition method to aid in mental calculations.

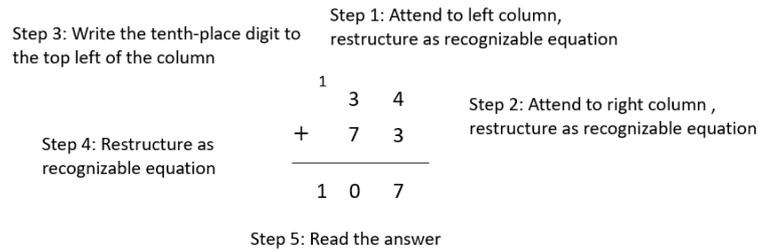

**Figure 6.2:** An illustration of the *addition method* to evaluate expressions with an infinite number of combinations. Endogenous attention is sequentially directed to specific regions of visual imagery, restructuring the arrival of percepts to match consolidated temporal sequences. **(1)** Attending to elements of the right column of the imagery in a sequential manner results in the perception of <4, +, 3>. Since <<4, +, 3>: <7>> is a common temporal sequence and is likely consolidated, the matching of temporal percepts leads to the activation of <7>. **(2)** Thereafter, attending to the left column leads to the matching of temporal sequence <<3, +, 7, >: <10>>, activating <10>. **(3)** Understanding that the sum has exceeded 10, the FFA executes an internal action that introduces the visual imagery of <1> to the upper-left column of the current imagery. Perceiving and understanding the new imagery, another action is generated to bring <1> down to the answer row. **(4)** Attention is directed sequentially from left to right of the answer row, reading out the final answer of <107>. Besides using mental imagery, the problem can also be solved through external RT by evaluating it on paper.

During reading, the temporal sequence of words entering perception is controlled by sequentially attending to text from left to right. When reading a brochure, attention first targets the larger fonts and images that convey the main message or theme, with leading lines guiding the eye to important information. When viewing a static image, attention is sequentially directed to prioritize salient features.

## Examples for (2)

Certain problems require formulating sub-questions to resolve the main question. For example, consider the question <To what extent do protectionist policies benefit Singaporeans?>, which can be reformulated into the sub-questions of, <How does protectionism benefit Singaporean industries?>, <How does protectionism promote economic growth?>. Encountering <To what extent …> questions may also prompt the respondent to reformulate it into the sub-questions of, <Protectionism policies harm Singaporeans because …>, <Protectionism benefits Singaporeans because …>, to develop a balanced analysis.

Now consider the question <What did you eat for lunch yesterday?>. One may prompt themselves with related sub-questions, such as <What did I do yesterday?>, <Where was I yesterday?>, or, if it was a Tuesday, <What do I normally do on Tuesdays?>. Broader prompts like <What food do I like to eat>, <Am I on a diet?> may also be used until one question triggers the retrieval of the relevant episodic memory. External RT may also be used through deliberate perception of relevant cues, which may induce the retrieval of episodic memory, such as the supplementary percepts of <dining table>, food items in the refrigerator, and money spent on food. Similarly, when attempting to identify a familiar face, one may prompt themselves with, < Where have I seen this person before?>. A student deepening their understanding of a lecture may ask, <What does this statement imply?>, <How is this relevant to concept B?>, or <How can this solution be further optimized?>, a detective investigating a murder may prompt himself with, <what does the evidence suggest?>, or <what is the perpetrator's motive?>.

## Examples for (3)

Certain tasks require the mental construction or imagination of a specific scene. For example, <Tom is sitting on a stool which is under a tent, above a field of grass>, followed by the question of <Is the stool above or below the field of grass?> requires the individual to construct a mental model of the scene. Similarly, <a square is two spots to the left of a circle, the circle is three spots to the right of a star, and a triangle is adjacent to the square and the circle> requires the utilization of visuospatial imagery and the active maintenance of object order, to determine the arrangement of objects in this scene.

Likewise, for external RT, an individual who has never seen an umbrella may find it difficult to imagine what an umbrella looks like when it opens. Individuals who have never played the guitar before may not realize that tensioning of the strings affects the pitch when strummed. The only way to discover the outcome is to experiment and observe the visual or auditory result afterward.

# 6.2 Decision making

Decision making involves the comparison of multiple decision prospects, where the threshold at which a decision is confirmed is determined by interactions between the FFA and CBGT. Decision thresholds are shaped through numerous experiences involving decision making, where RPE generated during each experience optimize the percept contexts in which a decision is confirmed. Two examples are discussed.

1. Under time constraints, the percept context for decision confirmation may encode a short elapsed time, as prior learning experiences may have associated delayed decisions with punishment.
2. The percept context for decision confirmation may encode predicted reward magnitude, or self-defined assurances, <I will buy shampoo if free lotion is included>, possibly sustained in the relation store. Once again, decision-making reflects learned principles shaped by RPEs across many prior experiences.

Decision-making can be enhanced by considering different perspectives and factors that may influence decision outcomes, which is a form of RT. Thereafter, derived reward values are temporarily related to their corresponding decision prospects through the relational store.

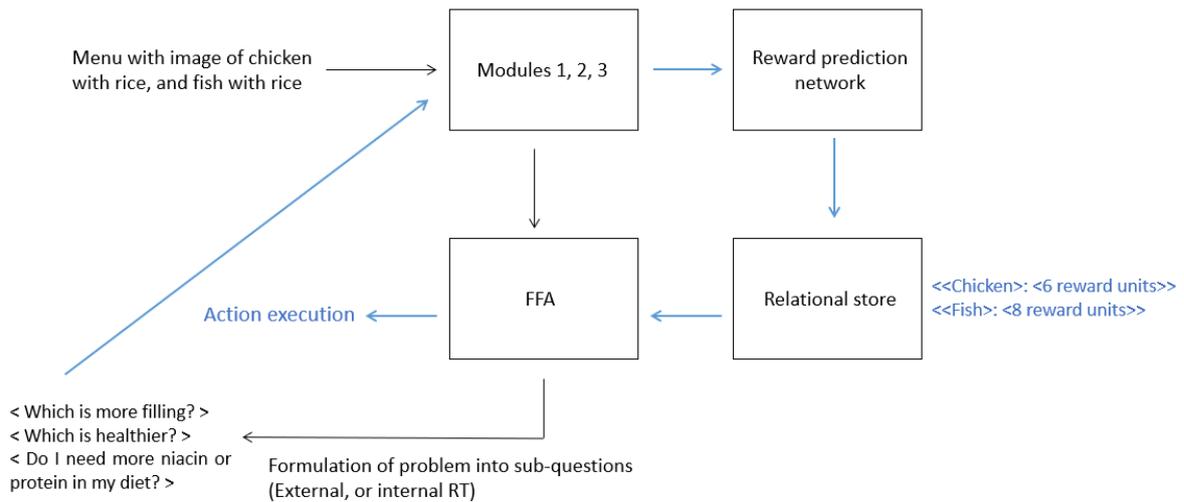

**Figure 6.3:** An illustration of decision making through the consideration of multiple factors for each decision prospect, and generating predicted reward values. Predicted reward values are related to their corresponding decision prospects in the relation store. Consider the scenario of being shown two pictures on the menu: one illustrating chicken with rice, while the other illustrates fish with rice. To decide which option to select, one may engage in visual or auditory RT, prompting themselves with the sub-questions: <Which is more filling, the fish or the chicken option?>, <Which is healthier?>, or <Do I need more DHAs in my diet?>. Thereafter, the predicted reward value of the decision prospect is then evaluated by the reward prediction network, with respect to each of these questions. Note that supplementary percepts encoding the anticipated time of the next meal, what was previously eaten, or whether DHA intake is recommended by dieticians, can influence the predicted reward value. Derived predicted values for each option can be temporarily stored and maintained in the relational store. Lastly, upon confirmation of the decision by the FFA, the appropriate course of action is executed.

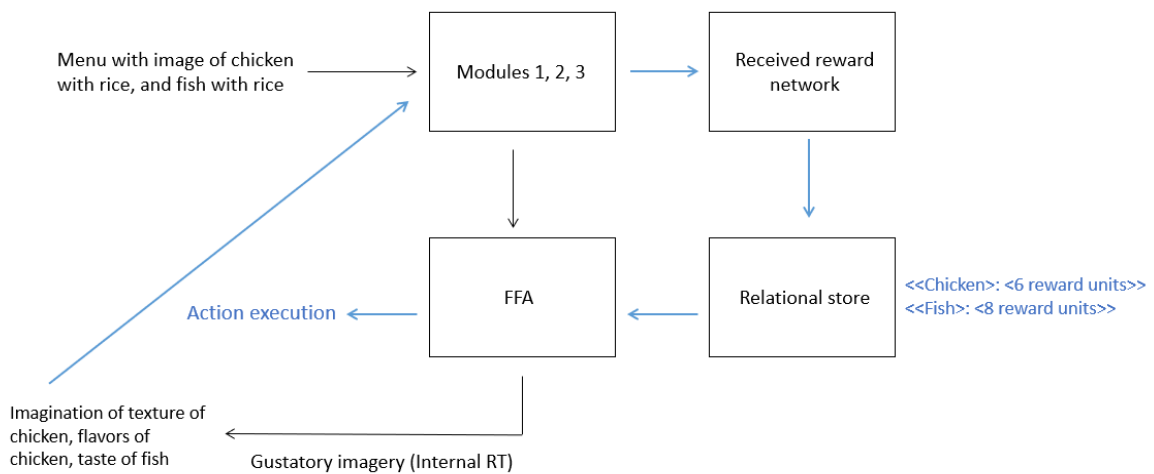

**Figure 6.4:** An illustration of decision making through consummatory imagination of received reward. Consider the scenario of being shown two pictures on a menu: one illustrating chicken with rice, while the other illustrates fish with rice. To decide which option to select, one may simulate the outcome of each decision prospect by imagining the taste, texture, and tenderness of chicken, and the flavors, deduced via the illustrations on the menu. The percepts derived

from imagery, coupled with other supplementary percepts, are input into the received reward network to generate a received reward value. Thereafter, the received reward values are related to their corresponding decision prospects in the relational store. Lastly, upon confirmation of the decision by the FFA, the appropriate course of action is executed.

## 6.3 Meta learning

Meta learning or "learning how to learn" emerges from trial-and-error learning guided by RPEs. When placed in some new environment, previously learned meta learning strategies can be applied to exploratory actions, such as directing eye movements in a specific manner, physically interacting with the environment in specific, or deliberately choosing what to learn, which may include directing attention towards salient percepts, suppressing and discarding certain percepts from working memory, or adjustment of priorities during certain situations. Meta-learning strategies provide modulation over a system that would otherwise indiscriminately and arbitrarily encode percept relations, failing to filter useful relations in new environments.

## 6.4 Formalization of operations

### <u>Operations</u>

$$P_{AE}, \ P_E, P_{ATL}, P_{CM}, P_S, P_M = P_{all}$$

From left to right: abstract executive percepts, executive percepts, ATL percepts, sensory percepts, and motor percepts. They are collectively regarded as $P_N$. Note that percepts can include both regular and temporal percepts.

$P_{pred}, P_{cue}, P_{com}$ refers to predicted percepts, the cue percept, and the command percept, respectively. $A_{seq}$ refers to an action sequence, which may encompass internal or external actions.

### <u>Update function</u>

Operation: Ungate (*Sender_region, Target_region, Information_to_update*)

The prerequisite for ungating is bidirectional availability: the target region must be available to receive signals, and the input region must be available to send signals. Ungating is facilitated by CBGT loops and inhibitory dynamics within regions.

### <u>Protect function</u>

Operation: Protect (*Region_protected_from_distractions*)

The protect function is the inverse of the ungate function, preserving the current information and preventing signal overwriting within a region.

### <u>Propagate function</u>

Operation: Propagate (*Region, Percepts_to_propagate*): *Percepts_activated*

The process of activating input percepts and deriving target percepts through relational propagation.

## Generation of an action sequence mediated by higher-order controller

Propagate (FFA, $P_N$) → $A_{seq}$ *// FFA initiation of internal action or external action*

## Retrieval of mental imagery (<cat (lex)> activates mental imagery of a cat)

Operation: Static_Imagery()

$A_{seq}$ → Update (PFC, Sensory areas, $P_N$) *// Activation of mental imagery*

Operation: Dynamic_Imagery()

Propagate (Hippocampus, $P_N$) → $P_{pred}$ *// Auto-associating of episodic memory*

$A_{seq}$ → Update (Hippocampus, Sensory areas, $P_{pred}$) *// Activation of mental imagery*

## Relational propagation

Operation: RP()

$A_{seq}$ → Update (PFC, ATL, $P_N$)

→ Propagate (ATL, $P_N$) → $P_{new}$

→ Propagate (TPS, $P_{ATL}$) → $P_{new}$

→ Update (TPS, PFC, $P_{new}$)

## Command/goal-oriented response

Operation: Rule_Internalization()

RT ()

$A_{seq}$ → Update (VLPFC, DLPFC, $P_N$) *// $P_N$ encompasses $P_{Cue}$*

$A_{seq}$ → Protect (DLPFC)

RT () *// Imagery of cue percept and command percept*

$A_{seq}$ → Update (VLPFC, DLPFC, $P_N$) *// $P_N$ encompasses percept command, temporarily relating the cue percept to the command percept*

$A_{seq}$ → Protect (DLPFC)

Propagate (DLPFC, $P_N$) → $P_{com}$ // Realization of percept context matching stored cue $P_N = P_{cue}$, outputting the command related to the cue

$A_{seq}$ → Propagate (FFA, $P_N + P_{com}$) // Activation of the goal

### Retrieval of episodic memory

RP()

$A_{seq}$ → RT ()

// Shuffling between the processes above

$A_{seq}$ → Imagery ()

$A_{seq}$ → Dynamic_Imagery ()

### Decision making between *n* decision prospects/cost-benefit analysis

RP()

$A_{seq}$ → RT () // Imagining possible factors of consideration

// Shuffling between the two processes for complex decision prospects involving limitations, logical implications, morals, etc.

$A_{seq}$ → Update (VmPFC, OFC, $P_N$) → $V_{all}$ // Predicted reward and received rewards

$A_{seq}$ → Update (VLPFC, DLPFC, $P_N + V_N$) // Storage of reward variable item

$A_{seq}$ → Protect (DLPFC)

// Repeat for other decision prospects

### Application of rule/context/goal/command, immediately/over time

Propagate (FFA, $P_N$) → $A_{seq}$ // Percept commands that demand immediate action will execute action without delay

### Working memory manipulation (reciting a sequence backward)

$A_{seq}$ → Update (VLPFC, DLPFC, $P_N$) // Storage of memory, repeat for the length of the sequence

$A_{seq}$ → Propagate (DLPFC, $T_n$) → $P_n$ // Activation of time cell and corresponding percepts

### "If A leads to B, C leads to B" type reasoning

*Combination of TPS(), and ATL()*

### Prospective memory (situational)

Rule_Internalization()

Rule_Activation()

## Planning for goals, events, the future, and actions

RP ()

$A_{seq} \rightarrow$ RT () // Imagining possible factors of consideration

// Shuffling between processes above

$A_{seq} \rightarrow$ Update (VmPFC, OFC, $P_N$) $\rightarrow R_{all}$ // Predicted or received rewards

$A_{seq} \rightarrow$ Update (VLPFC, DLPFC, $P_N$)

## General question solving

RP ()

$A_{seq} \rightarrow$ RT () // Paraphrasal, reformulation, chain-of-thought

// Shuffling between processes above

$A_{seq} \rightarrow$ Update (VLPFC, DLPFC, $P_N$)

## Flexible rule switching and strategy selection

$A_{seq} \rightarrow$ Update (VLPFC, DLPFC, $P_{all}$)

## Error detection and correction/self-monitoring

*Require conflict detection by the anterior cingulate gyrus (ACC) (not discussed)*

## Prospective memory (temporal)

$A_{seq} \rightarrow$ RT ()

$A_{seq} \rightarrow$ Update (VLPFC, rPFC, $P_N$)

$A_{seq} \rightarrow$ Protect (rPFC) // Storage

$A_{seq} \rightarrow$ RT ()

$A_{seq} \rightarrow$ Update (VLPFC, rPFC, $P_N$) // Retrieval

Propagate (rPFC, $P_N$) $\rightarrow P_{com}$ // $P_N$ may encompass abstract executive percepts that can encode elapsed time

$A_{seq} \rightarrow$ Propagate (FFA, $P_N + P_{com}$)

## Filtering distractions

$A_{seq} \rightarrow$ Protect (DLPFC)

## Metacognition, confidence, and familiarity, judgment

*Abstract executive percepts*

**Habit suppression**

*Facilitated by CBGT and ACC*

**Multi-tasking**

*A combination of event prospective memory and temporal prospective memory*

**Visuospatial manipulation**

*Dynamic visual imagery*

# 7. Conclusion

This paper has presented a biologically feasible and functionally extensive model of the human brain, from the arrival of external stimuli through modality-specific processing, integration into abstract conceptual information, and influence on behavior. Key aspects of cognition, involving memory, learning, problem-solving, motivation, and decision-making, have been accommodated by the model proposed and substantiated through multiple real-world examples. However, there are limitations. Inconsistencies with experimental evidence may exist due to functional requirements, and oversimplified modules may lack detail and biophysical nuance observed in a real brain. The brain is an incredibly complex structure, with hundreds of specialized areas that process different stimuli. It is impossible to model every specialized area to achieve a fully human-like architecture.

Critically, this model proposes a different approach to achieving AGI, namely a modular design integrating multiple specialized functions within a hierarchical framework, with modules that can interact dynamically with one another. Instead of a giant monolithic decoder network in current LLMs, a model that shares functional and structural similarities to the human brain might be the key to understanding AGI.